\documentclass[10pt,twocolumn,letterpaper]{article}

\usepackage[english]{babel}
\usepackage{booktabs}
\usepackage{graphicx}
\usepackage[table,xcdraw]{xcolor}

\usepackage{cvpr}              % To produce the CAMERA-READY version
\usepackage{comment}
\usepackage{multirow}
\usepackage{makecell}
\usepackage{overpic}
\usepackage{float}

\usepackage{caption}       % for \captionof
\usepackage{cuted} %< for strip environment

\usepackage{currfile} %< for \currfilebase macro

\usepackage{caption} %< for "\captionof" commands (teaser)
\newcommand{\at}[1]{{\color[rgb]{0,0,0} #1}}

\usepackage{soul}
\setuldepth{foobar}

\renewcommand{\paragraph}[1]{\vspace{.5em}\noindent\textbf{#1.}}

\definecolor{cvprblue}{rgb}{0.21,0.49,0.74}
\usepackage[pagebackref,breaklinks,colorlinks,allcolors=cvprblue]{hyperref}

\def\paperID{3811} % *** Enter the Paper ID here
\def\confName{CVPR}
\def\confYear{2026}

\title{{\LARGE Spherical Voronoi} \\ Directional Appearance as a Differentiable Partition of the Sphere}

\author{
Francesco Di Sario$^{1,2}$ \qquad
Daniel Rebain$^{3}$ \qquad
Dor Verbin$^{5}$ \\
Marco Grangetto$^{1}$ \qquad
Andrea Tagliasacchi$^{2,4}$ \\
\\
$^1$University of Torino \qquad
$^2$Simon Fraser University \qquad
$^3$University of British Columbia \\
$^4$University of Toronto \qquad
$^5$Google DeepMind
}

\begin{document}
\maketitle
\begin{strip}
  \vskip -4.7em
  \centering
  \includegraphics[width=0.9\textwidth]{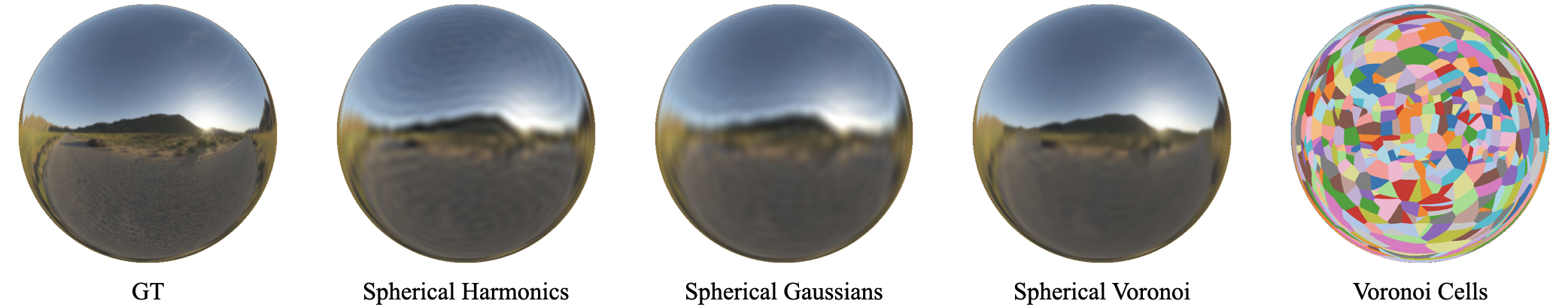}
  \captionof{figure}{
    Spherical functions, such as the shown environment maps, are widely 
    used in Computer Graphics and 3D Computer Vision. Classical 
    representations like Spherical Harmonics optimize well but struggle 
    with high frequencies. Explicit representations like Spherical 
    Gaussians capture localized functions, but are hard to optimize due 
    to kernel locality. We propose Spherical Voronoi, an explicit 
    representation that models \textit{high frequencies} effectively, 
    provides an \textit{adaptive} spherical-domain decomposition, and 
    is \textit{easier to optimize}.}
  \label{fig:\currfilebase}
\end{strip}
\begin{abstract}
Radiance field methods (e.g.~3D Gaussian Splatting) have emerged as a powerful paradigm for novel view synthesis, yet their appearance modeling often relies on Spherical Harmonics (SH), which impose fundamental limitations. SH struggle with high-frequency signals, exhibit Gibbs ringing artifacts, and critically fail to capture specular reflections. While alternatives like Spherical Gaussians offer improvements, they introduce significant optimization complexity. We propose Spherical Voronoi (SV) as a unified framework for appearance representation in 3D Gaussian Splatting. SV partitions the directional domain into learnable regions with smooth boundaries, providing an intuitive and stable parameterization for view-dependent effects. For diffuse appearance, SV achieves competitive results while maintaining simpler optimization compared to existing alternatives. For reflections --- where SH fundamentally fail --- we leverage SV as learnable reflection probes, taking reflected directions as input following principles from traditional graphics. This formulation achieves state-of-the-art results across both synthetic and real-world datasets, demonstrating that SV offers a principled, efficient, and general solution for appearance modeling in explicit 3D representations. 
Project page: \href{http://sphericalvoronoi.github.io}{sphericalvoronoi.github.io}
\end{abstract}
    
\section{Introduction}
\label{sec:intro}
In recent years, novel view synthesis has seen remarkable progress, 
driven largely by radiance field-based methods. While 
current state-of-the-art approaches effectively capture fine-grained 
texture and intricate geometric detail, they struggle to reproduce 
the complex, view-dependent appearance inherent to glossy surfaces.
This fact, along with the plateau in photometric accuracy at the top 
of novel view synthesis benchmarks~\cite{nerfbaselines}, clearly 
indicates that further improvement will require fundamental changes 
in how the \textit{directional} component of radiance fields is modeled. The view-dependent appearance representation of NeRF~\cite{nerf} 
employs a large MLP to model emitted radiance at each 3D point and 
viewing direction, achieving high-fidelity novel view synthesis. 
However, this formulation memorizes appearance directly from 
observations without explicitly reasoning about light transport, 
making it ill-suited for modeling specular reflections. Inspired by 
classical graphics, methods like Ref-NeRF~\cite{refnerf} address 
this by introducing more principled models of light transport. Such 
models take advantage of the fact that incident radiance changes 
slowly with position: when illumination originates sufficiently far 
from the scene, modeling reflections via a spherical environment map 
becomes extremely effective~\cite{physg}. More complex approaches 
additionally account for near-field light sources and 
inter-reflections via secondary ray tracing~\cite{nerfcasting, 
3dgrt, wu2024neural}, or jointly optimize geometry and 
illumination~\cite{mai2023neural, verbin2024eclipse, 
hasselgren2022nvdiffrecmc, Munkberg_2022_CVPR}, achieving 
high-quality results at the cost of significant computational 
complexity.

\begin{figure}
\centering
\includegraphics[width=0.9\linewidth]{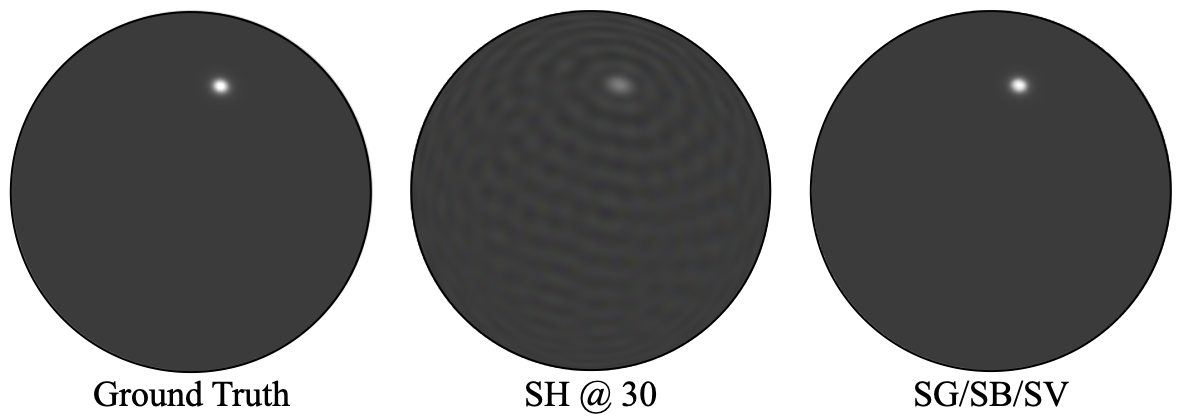}
\caption{\textbf{Gibbs artifacts} --
\at{Representing a ``glint'' on a shiny sphere like shown here is computationally impractical for the popular Spherical Harmonics representation.
Many coefficients (30 here) have to be used to reconstruct high-frequency functions accurately, and when this happens, Gibbs artifacts appear.
}
}
\label{fig:gibbs}
\end{figure}

Modern radiance field methods, such as Gaussian 
splatting~\cite{gsplat} and other fast 
approaches~\cite{plenoxels,radfoam,ever}, represent scenes as 
collections of explicit primitives rendered via fast rasterization. 
Each primitive stores view-dependent appearance using low-order 
spherical harmonics (SH, typically degree 3). SH are cheap to 
evaluate and easy to optimize thanks to the smoothness they induce 
in the loss landscape and their global support over the spherical 
domain. However, SH only allow band-limited reconstructions, and 
the number of coefficients required grows quadratically with 
frequency, making accurate modeling of sharp specular highlights 
impractical (see~\Cref{fig:gibbs}). Recent extensions of Gaussian 
splatting attempt to address this, but either assume far-field 
illumination only~\cite{3dgsdr}, or resort to implicit neural 
decoders~\cite{refgs}, limiting their physical accuracy or 
efficiency.

While bridging the gap between lighting decomposition and direct 
appearance memorization in complex scenes remains an open challenge, 
we take a step toward unifying these paradigms by proposing 
\emph{Spherical Voronoi} (SV), a new explicit representation for 
spherical functions inspired by differentiable Voronoi 
diagrams~\cite{voronoinet, derf, radfoam}. Our contributions are:

\begin{itemize}
    \item \textbf{Spherical Voronoi representation.}
    A new explicit representation for spherical functions based on a 
    differentiable soft Voronoi partition of the sphere, overcoming 
    the limits of common spherical bases.

    \item \textbf{Directional radiance modeling.}
    Application of SV to view-dependent radiance in Gaussian Splatting pipelines, consistently improving reconstruction quality over standard spherical bases across multiple benchmarks.

    \item \textbf{Reflection modeling.}
    Extension of SV to spatially varying reflections, enabling a fully 
    explicit and differentiable appearance model that achieves 
    state-of-the-art results on reflective benchmarks.
\end{itemize}
\begin{figure}[t]
\centering
\includegraphics[width=0.9\linewidth]{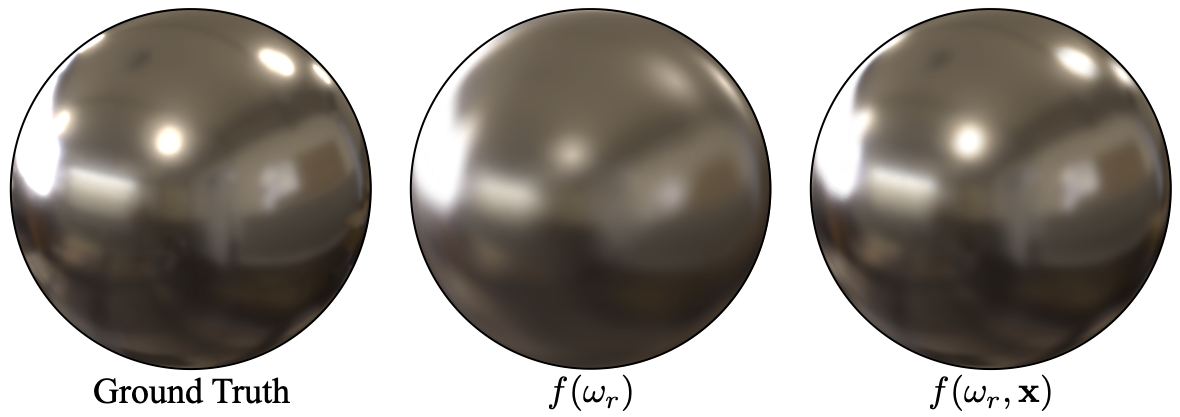}
\caption{\textbf{Spatially varying radiance} --
When illumination is spatially varying but a spatially-invariant model is used, conflicting measurements are averaged, resulting in a blurry reconstruction (middle).
Conversely, when a spatially varying model is used, crisp functions can be recovered.
}
\label{fig:\currfilebase}
\end{figure}

\begin{figure*}
\centering
\includegraphics[width=0.9\linewidth]{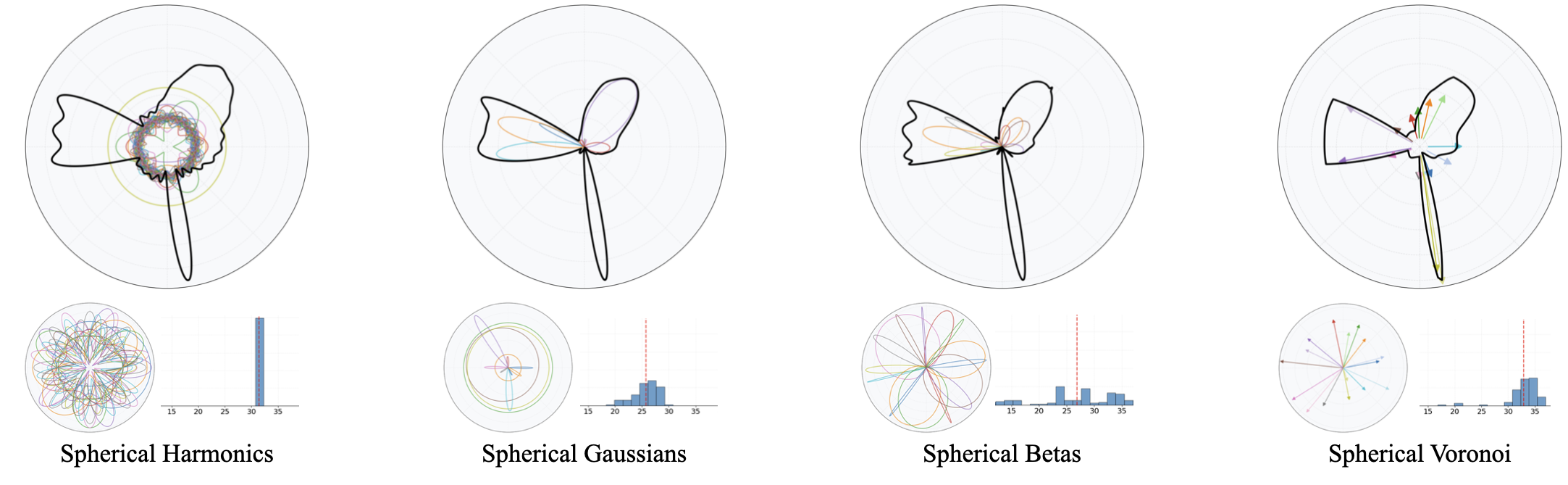}
% \begin{tabular}{*{4}{p{.23\linewidth}}}
%     \centering Spherical Harmonics&
%     \centering Spherical Gaussians&
%     \centering Spherical Beta&
%     \centering Spherical Voronoi
% \end{tabular}
\caption{\textbf{Fitting functions on spherical domains (2D)} -- 
\at{We fit different spherical representations to the same functions starting from random initializations~(samples shown in the bottom-left plots).
To measure the robustness of optimization we repeat the fitting $100$ times from random initializations, and report the PSNR attained at the end of optimization (bottom-right plots).}
\at{Spherical Harmonics converge to a unique solution due to the orthogonality of its basis functions, but its band-limited characteristics result in Gibbs fitting artifacts.}
\at{Spherical Gaussian and Betas can represent locally supported functions, as well as discontinuities, but their optimization is not stable and prone to local minima.
\at{Spherical Voronoi \textit{consistently} converges to a \textit{better} reconstruction.
\at{We omit drawing the ground truth function, as it is practically identical to the one reconstructed by Spherical Voronoi.}
}
}
}
\label{fig:fitting2d}
\end{figure*}

\section{Related Works}
\label{sec:related}
NeRFs~\cite{nerf} represent scenes as continuous volumetric functions outputting density and view-dependent color, but originally required long training and struggled with specular surfaces. 
\citet{gsplat} replaced neural fields with millions of anisotropic Gaussians and fast rasterization, achieving real-time, high-quality view synthesis.

\paragraph{Modeling view-dependent effects}
NeRF's MLP-based view dependent appearance inspired alternatives that model view-dependent lighting more explicitly. \citet{plenoxels} replaces the MLP with a sparse voxel grid storing spherical harmonics (SH) per cell, achieving NeRF-quality results with a significant speed-up. In Gaussian splatting, early methods also used low-order SH, but richer bases have emerged.
\citet{radsplat} uses spherical Gaussians and a NeRF-trained prior for robust, ultra-fast optimization, while 
\citet{dbs} replaces Gaussians with bounded spherical Beta kernels, capturing sharper highlights without requiring normals.
\citet{scaffoldgs} uses anchor Gaussians and an MLP to generate view-dependent attributes per frame, reducing primitive count while accurately modeling reflections and transparency.

\paragraph{Modeling reflections (NeRF)}
Rendering specular reflections remains a major challenge for radiance fields.
\citet{refnerf} addresses this by factorizing radiance with surface properties and regularizing normals for smoother highlights.
\citet{nerfcasting} and \citet{wu2024neural} improve view-consistent reflections by tracing secondary rays through the field.
\citet{nero} jointly optimizes geometry and BRDF under estimated lighting, while \citet{ENVIDR} uses a learned neural renderer with an SDF-based model to simulate reflections and enable relighting.

\paragraph{Modeling reflections (3DGS)}
Gaussian splatting has been extended to handle reflections through methods, like GaussianShader~\cite{GaussianShader}, which adds a simplified shading model to each splat by storing diffuse and specular terms, along with a learned residual to capture higher-order effects. It estimates per-splat normals via ellipsoid axes and learned corrections, enforcing consistency with rendered geometry. This improves reflective realism while preserving real-time speed.
\cite{3dgsdr} complements this with deferred shading: splats are rendered to G-buffers, and a second pass computes reflections using estimated normals. By back-propagating reflection errors and refining normals across splats, it achieves sharper specular effects with minimal performance cost.
More recently, work has focused on factorizing lighting and reflectance in splatting.
\citet{3igs} introduces a learned local illumination field via low-rank tensor factorization and assigns each Gaussian a BRDF descriptor, enabling accurate view-dependent effects while keeping training and rendering efficient.
\citet{refgs} applies directional light factorization in screen space, combining a multi-scale spherical mip-map for roughness-aware reflection blur with a geometry-lighting factorization.
This achieves photorealistic, smooth reflections in complex scenes while maintaining interactive performance and accurate geometry.
\section{Method}
\label{sec:method}
We overview classical representations for spherical functions in~\Cref{sec:background}, and present our Spherical Voronoi representation in~\Cref{sec:sv}.
We then apply Spherical Voronoi to two core applications in radiance fields applications:
\begin{enumerate*}[label=(\roman*)]
\item fitting in a radiance field (\Cref{sec:radiance}), and 
\item modeling of reflections (\Cref{sec:reflections}).
\end{enumerate*}

\subsection{Background}
\label{sec:background}
A function defined on the sphere $f: \mathbb{S}^2 \rightarrow \mathbb{R}^C$ can be represented using a variety of spherical bases.
In what follows, and without loss of generality, we only consider scalar functions $C{=}1$.
To contextualize our discussion, we first review the representations that are typically found in the differentiable rendering literature.

% \TODO{we need a figure that visually compares SH vs. SG vs. SB vs. SV (same function, different representations)}

\paragraph{Spherical Harmonics (SH)}
A widely adopted choice is the Spherical Harmonics expansion: 
\begin{equation}
    f_\text{SH}(\omega ; c) = \sum^L_{l=0}\sum^l_{m=-l}c_{lm}Y_l^m(\omega),
    %\nonumber
\end{equation}
where $c_{lm} \in \mathbb{R}$ are optimizable coefficients and $Y_l^m$ are the (real) SH basis functions.
Optimizing SH is numerically stable: they form an orthonormal basis, and they are globally supported.
However, capturing sharply localized signals (\ie, high-frequencies) requires a large number of basis functions (large $L$), leading to both large parameter counts, and the introduction of Gibbs-like ringing artifact when representing discontinuities in the function; see \Cref{fig:gibbs} for an example.

\paragraph{Spherical Gaussians (SG)}
To better model localized signals, a linear combination of Spherical Gaussians can be employed:
\begin{equation}
    f_\text{SG}({\omega} ; \tau,s, c) = \sum_{k=1}^K c_k\exp \big( \tau_k(s_k \cdot \omega - 1)\big),
    %\nonumber
\end{equation}
where each lobe is defined by a mean direction $s_k$, concentration $\tau_k$, and amplitude $c_k$.
SGs describe the function as a combination of smooth and rotationally symmetric lobes.
However, as Gaussians (in practice) are compactly supported, the mixture of SG lobes is often sensitive to initialization, and unstable to optimize; see the histograms in~\Cref{fig:fitting2d}.
This leads to weak gradients when a lobe is misaligned, and to ill-conditioned updates for large concentration parameters $\tau_k$.

\paragraph{Spherical Betas (SB)} 
Spherical Betas extend SG by enabling asymmetric and bounded-support lobes defined as:
\begin{equation}
f_\text{SB}(\omega ; \alpha, \beta, s)
= \sum_{k=1}^K(1+s_k \cdot \omega)^{\alpha_k-1}\,(1-s_k \cdot \omega)^{\beta_k-1},
%\nonumber
\end{equation}
where $s_k$ is the principal direction and $\alpha_k, \beta_k{>}0$ control the shape.
These are more flexible than SGs, as they can model skewed and sharper directional variation.
However, this flexibility also makes them harder to optimize in practice.
Their contribution remains local, and extreme values of $\alpha_k$ or $\beta_k$ generate very sharp and flat regions that cause ill-conditioned gradients, and even stronger sensitivity to initialization than spherical Gaussians.

\begin{figure}[t]
\centering
\includegraphics[width=0.9\linewidth]{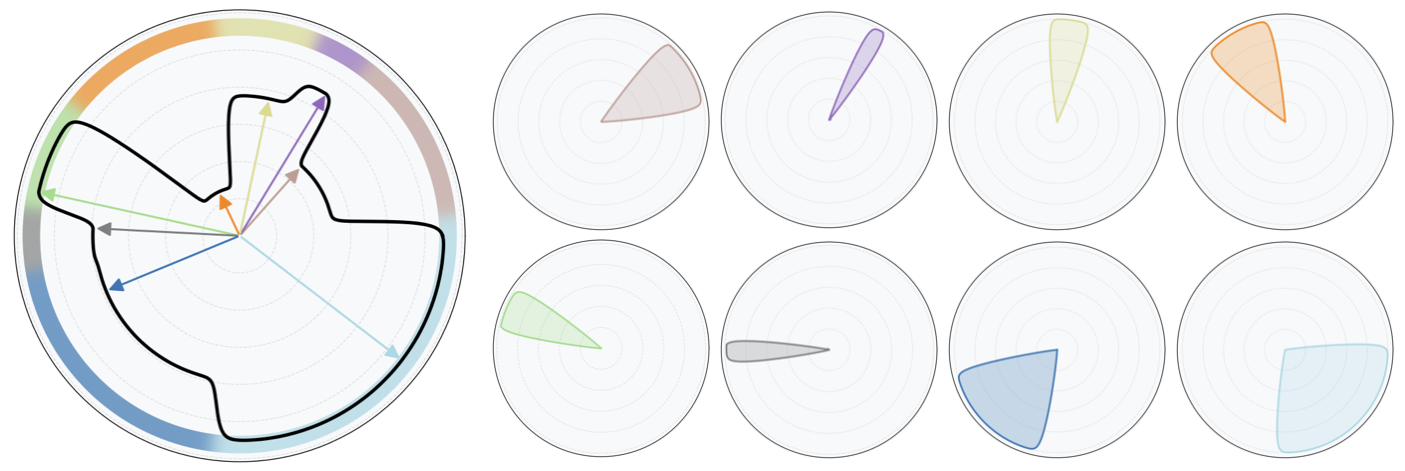}
\caption{\textbf{Spherical Voronoi Bases} -- 
\at{(left) a spherical function is represented by a collection of Voronoi sites.
(right) we visualize the support associated with each Voronoi site, and highlight how when combined they span the entire domain.}
% highlighting the fact that this support is a function of the proximity of other sites.
% The purple basis has a narrow support due to the proximity of the brown and yellow site, while the light blue basis spans a big section of the circular domain as no other sites are nearby.
% Notice that, together, , and therefore, during optimization, gradients are always able to propagate to the representation.
}
\label{fig:voronoi_2d}
\end{figure}

\subsection{Spherical Voronoi (SV)}
\label{sec:sv}
\at{To avoid these limitations, we introduce a new (\textit{soft}) Spherical Voronoi representation.
Our representation consists of a set of directional sites $s_1, ..., s_K \in  \mathbb{S}^2$ and associated function values $c_1, ..., c_K \in  \mathbb{R}$.
The function evaluation in direction $\omega$ is obtained as a weighted combination of the per-site values:}
\begin{equation} \label{eq:sv}
    f_\text{SV}(\omega ; \tau, s, c) = \sum_{k=1}^K w_k(\omega; \tau_k) c_k,
\end{equation}
where the weight $w_k$ is computed using a softmax function:
\begin{equation}
    w_k(\omega ; \tau) = \frac{\exp(\tau_k s_k \cdot \omega)}
                       {\sum^K_{k'=1} \exp(\tau_{k'} s_{k'} \cdot \omega)}.
\end{equation}
\noindent
\begin{figure}[t]
\centering
\includegraphics[width=0.9\linewidth]{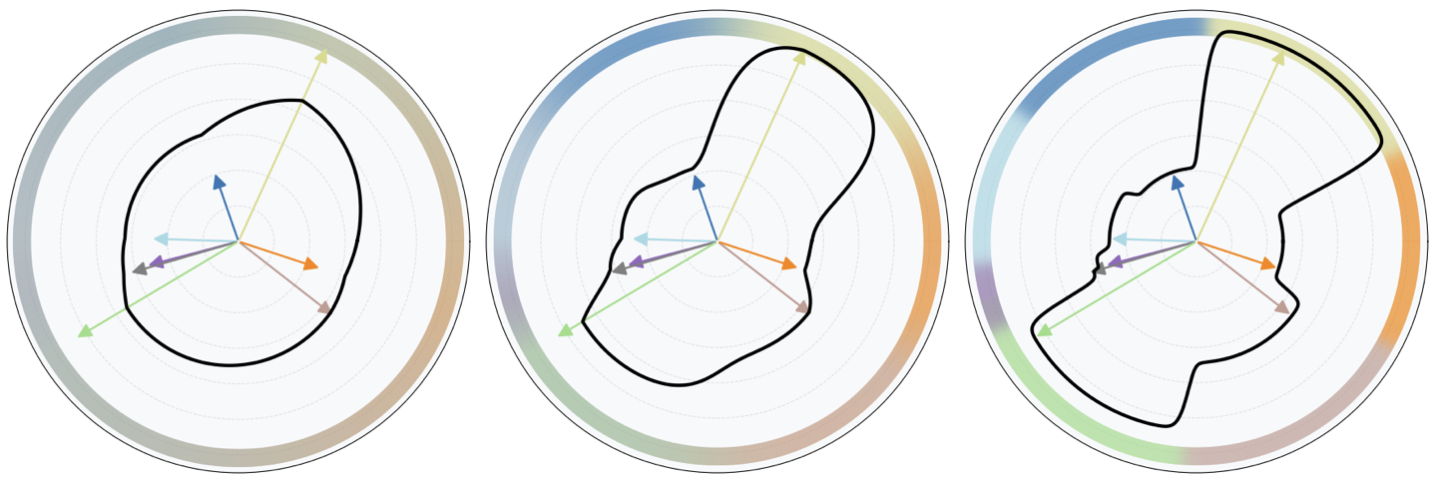}
\caption{\textbf{Effect of temperature $\tau$} --
\at{With a low value of~$\tau$ the representation is easy to optimize as its degrees of freedom have a large support, while if \textit{sharp} discontinuities are needed, a large value of $\tau$ can be chosen; above $\tau=\{1,5,25\}$.}
}
\label{fig:\currfilebase}
\end{figure}

% \TODO{$\downarrow$}
% – \dor{remove angles from outside}
% - \dor{remove numbers from inside}
% - \AT: tag the temperature you use
% 
% \endinput
% \dor{beautiful plot! I'd maybe remove the angles from the outside (and maybe also the numbers on the inside)---they're hard to read and don't add much.}
% \AT{+1}
% \AT{bit small... what about a 2x2 layout? or a two column}
\begin{figure*}[t]
\centering
\includegraphics[width=\linewidth]{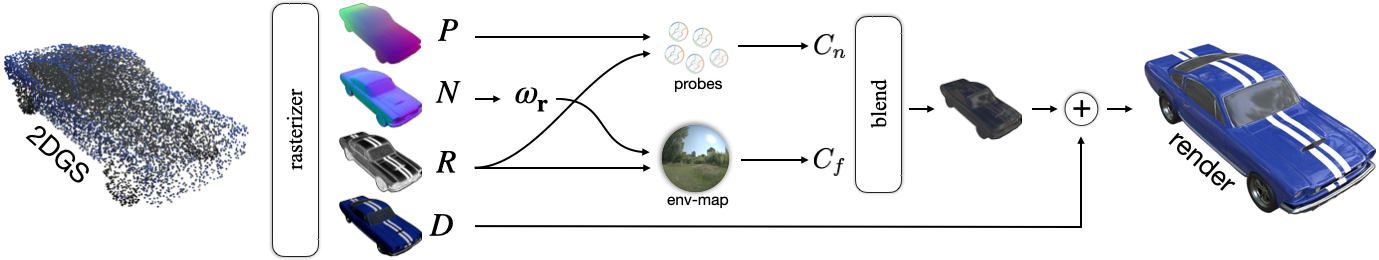}
% \includegraphics[width=\linewidth,height=.2\linewidth]{example-image-golden}
% \\
% \includegraphics[width=.195\linewidth]{example-image-golden}
% \includegraphics[width=.195\linewidth]{example-image-golden}
% \includegraphics[width=.195\linewidth]{example-image-golden}
% \includegraphics[width=.195\linewidth]{example-image-golden}
% \includegraphics[width=.195\linewidth]{example-image-golden}
% 
\caption{\textbf{Deferred rendering pipeline} -- The 2DGS scene is rasterized into buffers of (positions, normals, roughness, diffuse color).
These are then used in the lighting pass, combining light-probe illumination and an environment cubemap to compute diffuse and specular shading, which are blended to produce the final output.
}
\label{fig:def_pipeline}
\end{figure*}
The \textit{temperature} parameters $\tau_k {>} 0$ control the sharpness of the partition: small values produce smooth color transitions, while large ones approach a hard Voronoi tessellation, as illustrated in~\Cref{fig:temperature_effect}. When all $\tau_k$ share the same value, the formulation corresponds to the standard soft Spherical Voronoi model, whereas assigning distinct temperatures to each site naturally extends it to a weighted variant with locally adaptive angular sharpness.
By adjusting $\tau$, the model can effectively represent both smooth and sharp signals, typical in directionally localized structures.
This formulation works well in practice because the softmax ensures well-defined gradients for all sites, the temperature provides a way to adjust the sharpness of the representation, and the resulting soft partition induces a clean, non-overlapping decomposition of the sphere that avoids the ``competition'' between overlapping kernels which representations like SG and SB exhibit. Please refer to~\Cref{fig:fitting2d} for an analysis of approximation and convergence behavior.

\subsection{View-direction parameterization}
\label{sec:radiance}
\at{For classical radiance fields training, we replace the view-directional representation with our Spherical Voronoi.
Given a view-direction~$\omega$, the radiance associated with the primitive is evaluated as~$f_\text{SV}(\omega {;}\tau, s, c):\mathbb{S}^2 \rightarrow \mathbb{R}^3$, as defined in \Cref{eq:sv}, where the co-domain of the function is the RGB view-dependent color.
For 3DGS, that means that \textit{each} Gaussian will be equipped with additional parameters $\big(\{\tau_k\}, \{s_k\}, \{c_k\}, \big)$, and all of these parameters are treated as learnable and jointly optimized.}

\subsection{Reflection-based Parameterization}
\label{sec:reflections}
\at{\citet{refnerf} showed how, for glossy surfaces, when the view direction is used to query emitted radiance, a complex function ought to be learned from (relatively) sparse measurements.
Instead, they propose to co-learn the normals $n$ of surfaces, and reason in terms of a simpler function -- the radiance measured along the \textit{reflected} view direction $\omega_r = 2 \, (\omega \cdot n) \, n - \omega$}.
% 
% \begin{equation}
%     \omega_r = 2 \, (\omega \cdot \mathbf{n}) \, \mathbf{n} - \omega.
% \end{equation}
%
\at{Spherical Voronoi functions are sufficiently expressive to model highly complex directional behaviors, including multi-modal and discontinuous functions, and particularly well-suited to represent sharp illumination lobes (\eg, a specular highlight / glint on a surface).
This makes them appropriate not only for representing smooth view-dependent effects~$f(\omega)$, but also for learning fine-grained specular reflections~$f(\omega_r)$ in the spirit of~\citet{refnerf}.}

\paragraph{Spatially varying incoming radiance}
\at{Relying exclusively on $f(\omega_r)$ implicitly assumes \textit{far-field} illumination, which, in many real-world scenarios, is an incorrect model.
This becomes a problem when a glossy object is in proximity of other objects or light sources, and the appearance becomes a function of not just direction, but also position; see~\Cref{fig:spatially_varying_light}.}
\at{To account for spatial changes in the reflected light field, \citet{refnerf} conditioned the neural field on \textit{both} reflected direction and position.
However, when we parameterize the 3D scene with explicit representations like Gaussian splats, this becomes much more difficult to achieve (vs. conditioning by concatenation in neural fields).}

\paragraph{Learnable light probes}
To overcome this limitation, we introduce \emph{learnable light probes},  a set of spatially distributed, jointly optimized probes placed throughout the scene, each encoding a local reflection field as an explicit Spherical Voronoi function queried in the \emph{reflected} view direction.
Each probe stores a compact representation of the incoming radiance, enabling the model to approximate spatially varying lighting by interpolating contributions from nearby probes.
Unlike previous work~\cite{you2024nelf, xie2025envgs}, our probes encode an explicit directional appearance function queried in the reflected view direction, making them directly suited for specular appearance modeling in 3DGS.

\paragraph{Deferred Rendering with Voronoi Light Probes}
Following~\cite{3dgsdr, refgs}, we adopt a deferred rendering strategy that separates geometry from illumination. We build our framework upon the 2DGS backbone~\cite{2dsplat}, and extend each primitive  with two additional learnable material parameters: a roughness value $r \in [0,1]$ and a diffuse color $d \in \mathbb{R}^3$. 
In the \emph{geometry pass}, all 2D Gaussian primitives are rasterized once to produce per-pixel buffers at image coordinates $(u,v)$, storing the 3D position $P(u,v)$ of the visible surface, its normal $N(u,v)$, roughness $R(u,v)$, and diffuse color $D(u,v)$. In what follows all terms are computed per-pixel, so we omit the explicit dependence on $(u,v)$. %, assuming all terms are computed per-pixel.
The final shaded color for a given pixel is computed as:
\begin{equation}
C = D + C_{\text{spec}},
\end{equation}
where $C_{\text{spec}}$ encodes the specular component of the appearance. 
% The reflected view direction is defined as:
% \begin{equation}
% \omega_r = 2(\omega \!\cdot\! N)\,N - \omega.
% \end{equation}
The specular color is modeled as a spatially-varying blend between near- and far-field illumination:
\begin{equation}
\label{eq:c_spec}
C_{\text{spec}} = \alpha\,C_n + (1 - \alpha)\,C_f. %(\omega_r).
\end{equation}

The far-field term $C_f$ represents distant illumination and it is implemented as a learnable cubemap evaluated at the reflection direction $\omega_r = 2(\omega\cdot N)N-\omega$.

The near-field term $C_n$ captures spatially varying reflections using a set of learnable \emph{Voronoi light probes}. 
Each probe $i$ is parameterized by a position $p_i \in \mathbb{R}^3$, a blending weight $\alpha_i \in [0,1]$, and the parameters of a Spherical Voronoi function $(\tau_i, s_i, c_i)$, as defined in \Cref{eq:sv}. All of these parameters are optimized during training. %$f_{\text{SV}}^i(\omega; \tau, s, c)$.
For a surface point $P$, we query its $k$-nearest probes:
\begin{equation}
\mathcal{N} = \text{kNN}(P),
\end{equation}
and compute normalized inverse-distance weights:
\begin{equation}
\tilde{w}_i = \frac{\|P-p_i\|^{-1}}{\sum_{j\in \mathcal{N}} \|P-p_j\|^{-1}}.
\end{equation}
The near-field color and blending factor are then evaluated as:
\begin{align}
C_n &= \sum_{i\in \mathcal{N}} \tilde{w}_i\, f_{\text{SV}}^i(\omega_r; \tau), &
\alpha &= \sum_{i\in \mathcal{N}} \tilde{w}_i\, \alpha_i.
\end{align}
\begin{table*}[t]
\centering
\includegraphics[width=\linewidth]{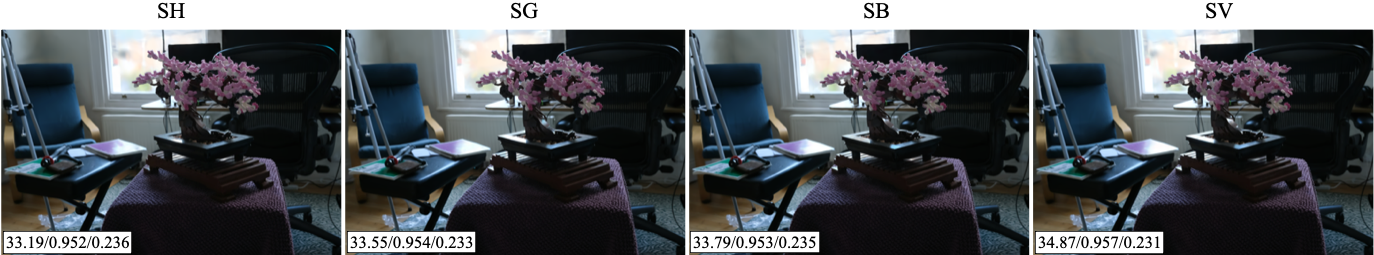}
\vspace{-1.5em}
% \small
\tabcolsep=0.08cm
\begin{center}
\resizebox{\textwidth}{!}{%
\begin{tabular}{lc|ccc|ccc|ccc|ccc}
                                 & \multicolumn{1}{l|}{}                                                                         & \multicolumn{3}{c|}{Mip-NeRF360}                                                              & \multicolumn{3}{c|}{NeRF-Synthetic}                                                           & \multicolumn{3}{c|}{DeepBlending}                                                             & \multicolumn{3}{c}{Tanks\&Temples}                                                            \\
\multirow{-2}{*}{Method}         & \multicolumn{1}{l|}{\multirow{-2}{*}{\begin{tabular}[c]{@{}l@{}}Color \\ Model\end{tabular}}} & PSNR $\uparrow$                          & SSIM $\uparrow$                          & LPIPS $\downarrow$                         & PSNR $\uparrow$                          & SSIM $\uparrow$                          & LPIPS $\downarrow$                         & PSNR $\uparrow$                          & SSIM $\uparrow$                          & LPIPS $\downarrow$                         & PSNR $\uparrow$                          & SSIM $\uparrow$                          & LPIPS $\downarrow$                         \\ \hline
Zip-NeRF~\cite{zipnerf}                         & MLP                                                                                           & \cellcolor[HTML]{FFCC99}28.55 & 0.829                         & \cellcolor[HTML]{FF9998}0.218 & 33.67                         & \cellcolor[HTML]{FF9998}0.973 & 0.036                         & -                             & -                             & -                             & 23.64                         & 0.839                         & \cellcolor[HTML]{FF9998}0.100 \\
                                 & SH                                                                                            & 28.09                         & \cellcolor[HTML]{FFCC99}0.833 & \cellcolor[HTML]{FFF8AD}0.231 & \cellcolor[HTML]{FFF8AD}34.15 & \cellcolor[HTML]{FFCC99}0.972 & \cellcolor[HTML]{FFCC99}0.033 & \cellcolor[HTML]{FFCC99}29.80 & \cellcolor[HTML]{FFCC99}0.912 & \cellcolor[HTML]{FFCC99}0.303 & \cellcolor[HTML]{FFF8AD}24.50 & 0.869                         & \cellcolor[HTML]{FFF8AD}0.175 \\
                                 & SG                                                                                            & \cellcolor[HTML]{FFF8AD}28.18 & \cellcolor[HTML]{FFCC99}0.833 & 0.233                         & \cellcolor[HTML]{FFCC99}34.26 & \cellcolor[HTML]{FFCC99}0.972 & \cellcolor[HTML]{FFCC99}0.033 & \cellcolor[HTML]{FFF8AD}29.67 & \cellcolor[HTML]{FFCC99}0.912 & \cellcolor[HTML]{FFF8AD}0.306 & 24.71                         & \cellcolor[HTML]{FFCC99}0.870 & 0.176                         \\
                                 & SB                                                                                            & 28.12                         & \cellcolor[HTML]{FFF8AD}0.831 & 0.238                         & 34.10                         & \cellcolor[HTML]{FFF8AD}0.971                         & \cellcolor[HTML]{FFF8AD}0.034 & 29.56                         & \cellcolor[HTML]{FFF8AD}0.907 & 0.316                         & \cellcolor[HTML]{FFCC99}24.54 & \cellcolor[HTML]{FFF8AD}0.866 & 0.196                         \\
\multirow{-4}{*}{Beta-Splatting~\cite{dbs}} & \textbf{SV}                                                                                   & \cellcolor[HTML]{FF9998}28.71 & \cellcolor[HTML]{FF9998}0.837 & \cellcolor[HTML]{FFCC99}0.228 & \cellcolor[HTML]{FF9998}34.58 & \cellcolor[HTML]{FF9998}0.973 & \cellcolor[HTML]{FF9998}0.032 & \cellcolor[HTML]{FF9998}30.63 & \cellcolor[HTML]{FF9998}0.917 & \cellcolor[HTML]{FF9998}0.298 & \cellcolor[HTML]{FF9998}25.00 & \cellcolor[HTML]{FF9998}0.874 & \cellcolor[HTML]{FFCC99}0.170
\end{tabular}%
}
\end{center}
% \vspace*{-1em}
\caption{\textbf{Modeling radiance} -- Our Spherical Voronoi model (SV) consistently improves performance over existing color parameterizations.
}
\label{tab:radiance_only}
\end{table*}

Here, the roughness $R$ modulates the temperature $\tau$, which controls the angular sharpness of the Spherical Voronoi distribution:
\begin{equation}
\tau = (1 - R)\, \tau_{\max} + R\, \tau_{\min},
\end{equation}
where $\tau_{\min}$ and $\tau_{\max}$ are fixed hyperparameters.
Lower roughness regions correspond to higher values of $\tau$, producing sharper reflections, while higher roughness broadens the lobes. 
This formulation provides a unified, differentiable and explicit model for diffuse and specular appearance. Note that in this reflection-based formulation, $\tau$ is not directly learned but derived from the surface roughness $R$. In contrast, in the radiance modeling setup of \Cref{sec:radiance}, $\tau$ is treated as a learnable parameter jointly optimized with the Spherical Voronoi sites and values.

\section{Experiments}
As mentioned in Section~\ref{sec:method}, we evaluate our approach in two settings. The first, which we refer to as \emph{view direction parameterization}, is discussed in \Cref{sec:exp_radiance} where we evaluate our Spherical Voronoi representation for modeling view-dependent appearance in ``standard'' representations which model outgoing radiance as a function of view direction (\ie, without explicitly modeling reflections). The second, which we refer to as \emph{reflection-based parameterization}, is presented in \Cref{sec:exp_reflections}, where we evaluate our Voronoi Light Probes representation by explicitly modeling reflections. In both settings, the Spherical Voronoi sites are initialized uniformly using Fibonacci sampling, and all remaining hyperparameters (losses, schedulers, \etc) follow the default configuration of the respective backbone.
% In both cases, we train for $30$K iterations following standard training procedures.

% \at{To re-iterate, we evaluate the two experimental configurations.
% In the first, we evaluate the use of Spherical Voronoi as a representation for view-dependent effects in radiance field workloads~(\cref{sec:exp_radiance}).
% In the second, we factor appearance into diffuse and specular components via Voronoi Light Probes~(\cref{sec:exp_reflections}).
% For both configurations, Spherical Voronoi sites are initialized uniformly using Fibonacci sampling, and all remaining hyperparameters (losses, schedulers, etc.) follow the default configuration of the respective backbones.
% In both cases, we train for $30$k iterations following the standard training procedure.}

\subsection{View Direction Parameterization}
\label{sec:exp_radiance}
%\footnote{\todo{We disable the early-stopping strategy introduced in the original method because it relies on test-set metrics, which we consider unfair.}
%\AT{avoid saying unfair, and explain it factually -- training with knowledge of the test set}
%}
%Instead, we re-evaluate all models on the already downsampled images to ensure a fair comparison.
% Additionally, as discussed in~\todocite{????}, on-the-fly sampling using PIL (as optionally available in the 3DGS codebase) avoids JPEG compression artifacts and improves quality by approximately $0.5$\,dB PSNR.
% Additionally, we compute reconstruction metrics without saving them as JPEG images, which avoids including compression errors in the quality metrics.
To evaluate our view-direction parameterization, we adopt the Beta Splatting backbone~\cite{dbs}. In their implementation, the model is periodically evaluated on the test set (every 500 iterations), and the best-performing checkpoint within 30k iterations is reported. Although this strategy effectively acts as an early-stopping mechanism, it relies on test-set feedback. In our experiments, we instead train for a fixed number of iterations without using the test set for model selection. 
In this setting, we employ the \textit{weighted} Spherical Voronoi parameterization introduced in Section~\ref{sec:sv}. Each site $s_k$ defines both a unit direction $\hat{s}_k = s_k / \|s_k\|$ and an implicit temperature $\tau_k = \|s_k\|$, with the norm of the site vector directly defining the temperature associated with that cell.
Additionally, for the \textit{MipNeRF-360} dataset we trained and tested the models \ul{on the already downsampled images}. Our Spherical Voronoi representation employs $8$ sites per Gaussian, matching the number of degrees of freedom of degree-3 spherical harmonics. The memory footprint scales linearly with the number of sites: each site stores three floats for its location and three for the radiance, while the temperature $\tau_k$ is computed on the fly from the site norm. This results in a total of 48 learnable parameters per Gaussian. We evaluate this configuration on standard radiance fields benchmarks: \textit{Mip-NeRF360}~\cite{mipnerf360}, \textit{DeepBlending}~\cite{deepblending}, \textit{Tanks\&Temples}~\cite{tanksntemples}, and \textit{NeRF-Synthetic}~\cite{nerf}.
\Cref{tab:radiance_only} shows the results of the view-direction parameterization experiments. Our Spherical Voronoi formulation yields consistent improvements over all baselines, achieving higher PSNR than all other color parameterizations (SH, SG, SB). Notably, it also surpasses Zip-NeRF, a strong neural radiance baseline \at{that has dominated benchmarks since its inception.}

\subsection{Reflection-based Parameterization}
\label{sec:exp_reflections}
\begin{figure*}[ht!]
\centering
\includegraphics[width=0.95\textwidth]{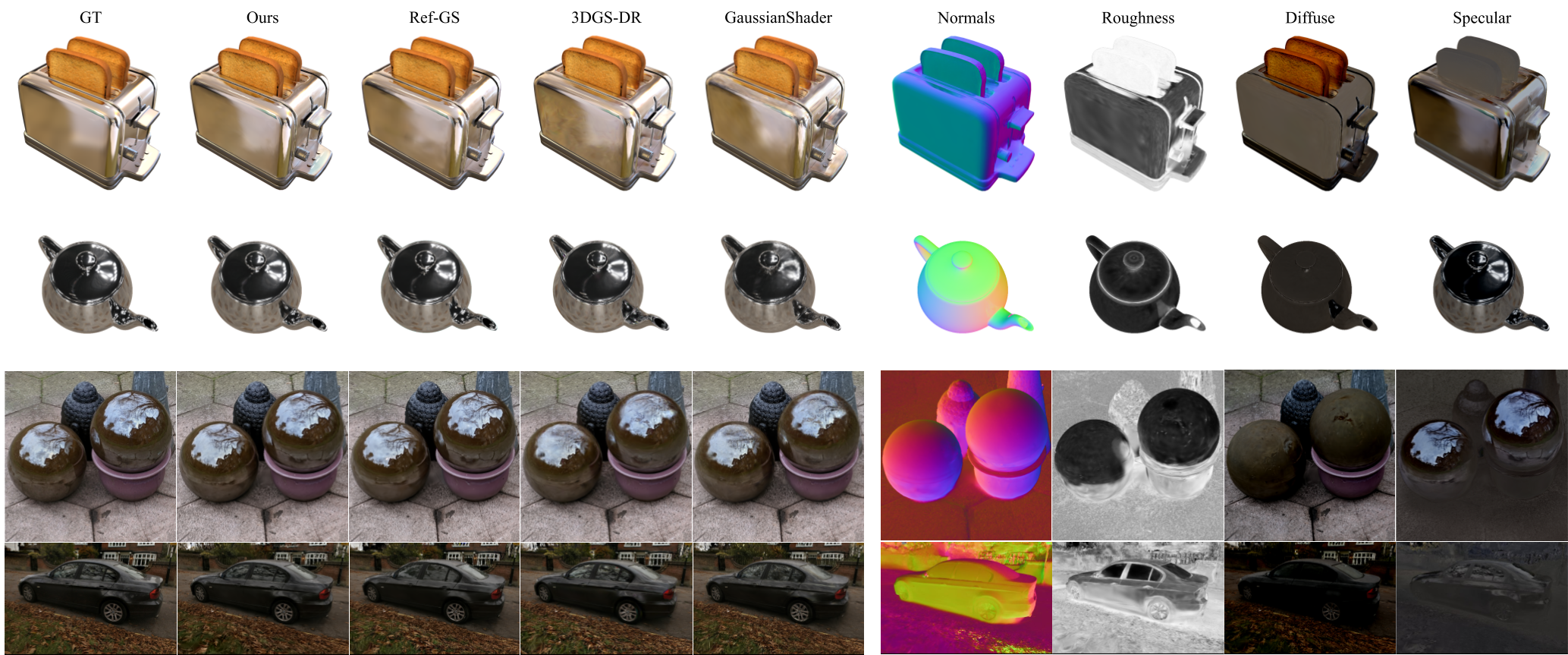}
\\[1em]
\resizebox{0.95\textwidth}{!}{
\begin{tabular}{lccc|ccc|ccc}
                         & \multicolumn{3}{c|}{Ref-NeRF~\cite{refnerf}}                                                                 & \multicolumn{3}{c|}{GlossySynthetic}                                                          & \multicolumn{3}{c}{Ref-Real}                                                                  \\
\multirow{-2}{*}{Method} & PSNR $\uparrow$                          & SSIM $\uparrow$                          & LPIPS $\downarrow$                         & PSNR $\uparrow$                          & SSIM $\uparrow$                          & LPIPS $\downarrow$                         & PSNR $\uparrow$                          & SSIM $\uparrow$                          & LPIPS $\downarrow$                         \\ \hline
Ref-NeRF~\cite{refnerf}                 & 32.32                         & 0.956                         & 0.110                         & 27.50                         & 0.927                         & 0.100                         & 23.62                         & 0.646                         & 0.239                         \\
NeRO~\cite{nero}                     & 29.84                         & 0.962                         & 0.072                         & -                             & -                             & -                             & -                             & -                             & -                             \\
ENVIDR~\cite{ENVIDR}                   & 32.88                         & 0.969                         & 0.072                         & 29.58                         & \cellcolor[HTML]{FFF8AD}0.952 & 0.057                         & 23.00                         & 0.606                         & 0.332                         \\
3DGS~\cite{gsplat}                     & 30.37                         & 0.947                         & 0.083                         & 26.50                         & 0.917                         & 0.092                         & \cellcolor[HTML]{FFCC99}23.85 & \cellcolor[HTML]{FFCC99}0.660 & \cellcolor[HTML]{FF9998}0.230 \\
GShader$^{\dagger}$~\cite{GaussianShader}           & 30.91                         & 0.954                         & 0.081                         & 27.54                         & 0.922                         & 0.087                         & 23.32                         & 0.648                         & 0.257                         \\
3iGS$^{\dagger}$~\cite{3igs}                   & 30.91                         & 0.950                         & 0.076                         & 26.86                         & 0.915                         & 0.088                         & 23.67                         & 0.644                         & \cellcolor[HTML]{FFCC99}0.232 \\
3DGS-DR$^{\dagger}$~\cite{3dgsdr}                & \cellcolor[HTML]{FFF8AD}34.13 & \cellcolor[HTML]{FFF8AD}0.972 & \cellcolor[HTML]{FFF8AD}0.058 & \cellcolor[HTML]{FFF8AD}30.36 & \cellcolor[HTML]{FFCC99}0.957 & \cellcolor[HTML]{FFF8AD}0.053 & \cellcolor[HTML]{FFF8AD}23.83 & \cellcolor[HTML]{FF9998}0.661 & 0.240                         \\
Ref-GS$^{\dagger}$~\cite{refgs}                & \cellcolor[HTML]{FFCC99}35.57 & \cellcolor[HTML]{FFCC99}0.975 & \cellcolor[HTML]{FFCC99}0.051 & \cellcolor[HTML]{FFCC99}31.27 & \cellcolor[HTML]{FF9998}0.962 & \cellcolor[HTML]{FF9998}0.045 & 23.81                         & \cellcolor[HTML]{FFF8AD}0.659 & \cellcolor[HTML]{FFF8AD}0.238 \\
\textbf{Ours}            & \cellcolor[HTML]{FF9998}36.09 & \cellcolor[HTML]{FF9998}0.976 & \cellcolor[HTML]{FF9998}0.050 & \cellcolor[HTML]{FF9998}31.30 & \cellcolor[HTML]{FF9998}0.962 & \cellcolor[HTML]{FFCC99}0.046 & \cellcolor[HTML]{FF9998}23.91 & \cellcolor[HTML]{FFF8AD}0.659 & 0.244                        
\end{tabular}%
}
\vspace{.5em}
\captionof{table}{\textbf{Modeling reflections} - Our method consistently achieves top-tier performance across all datasets. Methods marked with $^\dagger$ were retrained for consistency, and most results were successfully reproduced. Ref-GS performs slightly better than originally reported on Ref-NeRF and GlossySynthetic, but worse on Ref-Real, likely because the original implementation did not use the dataset’s pre-downsampled input images. In the top-left, we show example outputs: our method produces reflections that are generally on par with—if not superior to—Ref-GS, and clearly better than other baselines (e.g., sharper branch reflections in GardenSpheres and Sedan). On the right, we illustrate the learned decomposition of the scene in diffuse and specular color, roughness and normals.}

\label{tab:comparison-f}
\end{figure*}

For the reflection-based parameterization experiments, we use the 2DGS backbone~\cite{2dsplat}, which provides more accurate normal estimation.
For these experiments, we use $128$ probes for synthetic scenes and $1024$ probes for real scenes, combined with $2048$ sites. 
The cubemap resolution is $256 \times 256 \times 6 \times 3$. We evaluate this configuration on three widely used datasets for reflective scenes: \textit{Ref-NeRF}, \textit{GlossySynthetic}, and \textit{Ref-Real}.
Table~\ref{tab:comparison-f} presents results for the reflection-based parameterization experiments. Our method achieves state-of-the-art performance on \textit{Ref-NeRF} and \textit{GlossySynthetic} datasets, and obtains competitive metrics on \textit{Ref-Real}. Importantly, our method slightly outperforms \textit{Ref-GS}, which uses the same 2DGS backbone but relies on an MLP decoder to model near-field reflections, whereas our approach is fully explicit. 

% \subsection{Qualitative Results and Scene Decomposition}
% \begin{figure*}
%     \centering
%     \includegraphics[width=\linewidth]{figures/decomposition.pdf}
%     \caption{\FS{Would like to show also this}}
%     \label{fig:placeholder}
% \end{figure*}

\subsection{Ablations}
\begin{figure*}[t]
\centering
\includegraphics[width=0.9\linewidth]{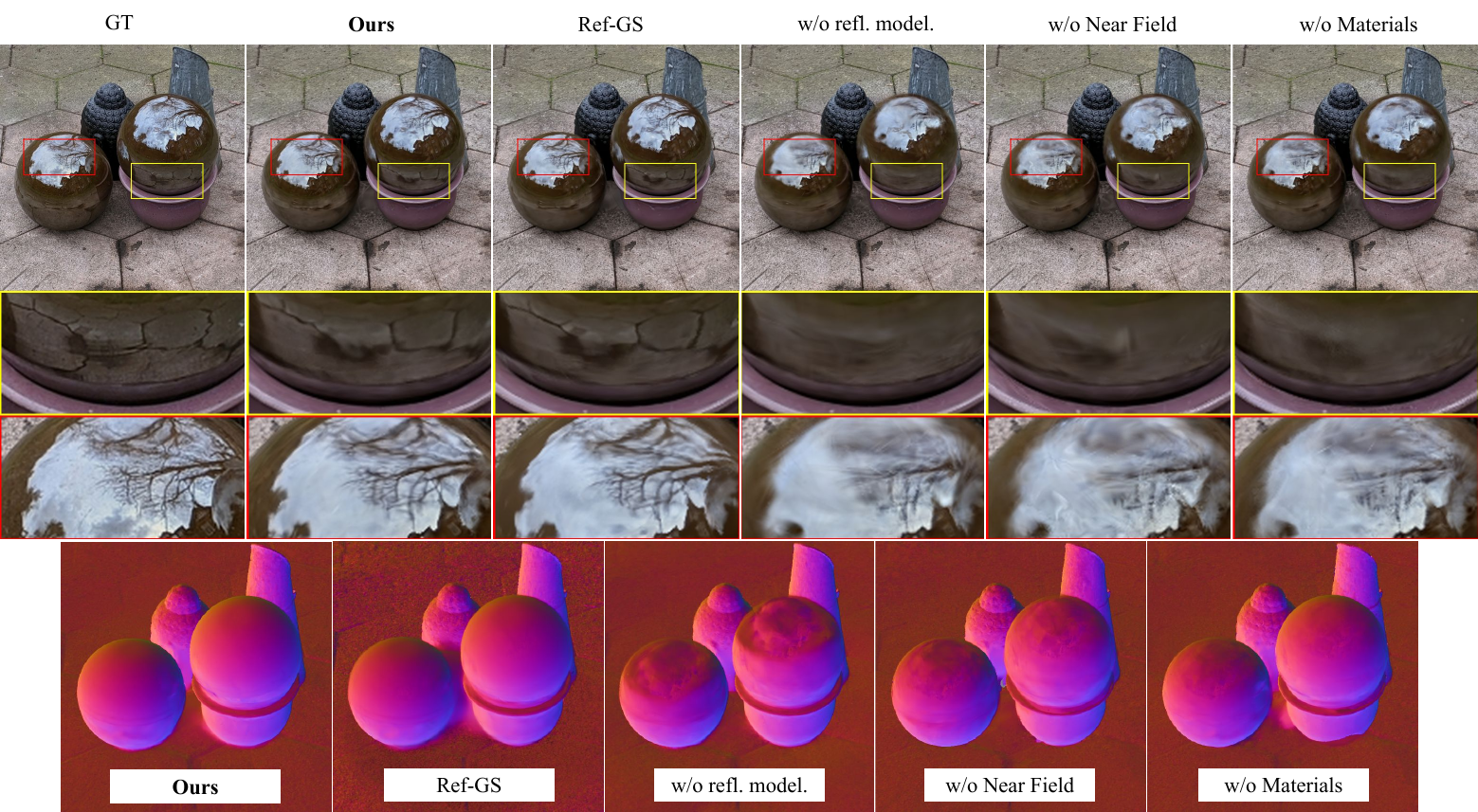}
\caption{\textbf{Qualitative ablation} --
We visualize the impact of each component of our model and compare against Ref-GS. Removing reflections, near-field probes, or material parameters (roughness) progressively reduces sharpness, reflection coherence, and normal quality, whereas our full SV-based formulation preserves high-frequency detail and stable shading.}
\label{fig:qualitative_ablation}
\end{figure*}

\begin{table}[t]
\begin{center}
\resizebox{.7\columnwidth}{!}{%
\begin{tabular}{lrrr}
        & \multicolumn{1}{c}{PSNR $\uparrow$}      & \multicolumn{1}{c}{SSIM $\uparrow$}      & \multicolumn{1}{c}{LPIPS $\downarrow$}     \\ \hline
SG     & 34.19                         & 0.967                         & 0.060                         \\
SB     & 34.07                         & 0.966                         & 0.061                         \\
Cubemap & 34.48                         & 0.970                         & 0.056                         \\
SV     & \cellcolor[HTML]{FF9998}36.09 & \cellcolor[HTML]{FF9998}0.976 & \cellcolor[HTML]{FF9998}0.050
\end{tabular}%
}
\end{center}
\vspace*{-1em}
\caption{\textbf{Modeling light probes} – Spherical Voronoi delivers substantially higher quality, evaluated on the Ref-NeRF dataset.}
\label{tab:lp_ablation}
\end{table}

\Cref{tab:lp_ablation} reports a quantitative comparison of alternative probe parameterizations used in our framework. We experimented with spherical Gaussians, spherical betas, and cube-map–based representations, keeping the number of learnable parameters \textit{fixed} across settings. Among all tested variants, the Voronoi-based representation consistently achieved the best reconstruction accuracy. This suggests that the inherent spatial partitioning induced by Voronoi cells provides a more expressive and compact basis for modelling local appearance, leading to significantly improved performance under identical capacity constraints.
Figure ~\ref{fig:qualitative_ablation} presents a qualitative ablation on the GardenSpheres scene, which is particularly challenging due to its complex geometry and high-frequency specular and local interactions. We visualize the contribution of each component of our model and compare the final reconstruction to Ref-GS, the current state-of-the-art. Despite relying entirely on an explicit representation, our approach produces reconstructions that are on par with—if not more detailed than—Ref-GS. Notably, our method better resolves fine structures such as tile patterns and other high-frequency textures, without requiring a neural model.
It is also important to highlight that standard perceptual and photometric metrics (PSNR, SSIM, LPIPS) remain comparable across all methods, even though models that explicitly model reflections tend to look qualitatively superior.

\subsection{Train and Inference Time}

\begin{table}[t]
\centering
\caption{\textbf{Efficiency comparison} - Rendering speed and training time reported as factors with respect to 3DGS.}
\label{tab:train_time}
\resizebox{0.75\columnwidth}{!}{%
\begin{tabular}{lr|r}
Method         & \multicolumn{1}{l|}{Rendering Speed} & \multicolumn{1}{l}{Train Time} \\ \hline
3DGS           & 1.00$\times$                                 & 1.00$\times$                          \\
GaussianShader & 0.17$\times$                               & 11.05$\times$                         \\
3iGS           & 0.44$\times$                                & 2.07$\times$                            \\
3DGS-DR        & 0.93$\times$                                & 3.25$\times$                            \\
Ref-GS         & 0.37$\times$                                  & 2.63$\times$                            \\
Ours           & 0.45$\times$                                  & 2.43$\times$                          
\end{tabular}%
}
\end{table}

For radiance-only modeling, both training and inference remain comparable to the original backbones, since using a small number of SV sites per Gaussian adds only a lightweight computation—essentially a dot product followed by a softmax. In this setting, our runtime closely matches SH-, SG-, and SB-based implementations.
The reflection-based model is naturally slower: deferred shading requires two rendering passes, and evaluating/interpolating the nearest light probes introduces an additional cost. Even with these overheads, our method runs at 0.45× the speed of 3DGS and is faster than Ref-GS, the current state-of-the-art for reflection modeling. Training time follows a similar pattern: probe optimization adds overhead, but our overall cost (2.43×) remains competitive and well below heavier formulations such as GaussianShader.

\section{Conclusion}
\label{sec:conclusion}
We introduced Spherical Voronoi functions as a new explicit representation for modeling directional appearance in radiance-field–based reconstruction. Thanks to their adaptive decomposition of the sphere and stable optimization behavior, SV consistently outperforms traditional bases such as SH, SG, and SB in radiance-only settings, while maintaining runtime comparable to the underlying forward-rendering backbones.
We further extended this representation to spatially varying reflections through Voronoi Light Probes, a fully explicit formulation that captures complex near-field effects without neural decoders. By relying on deferred shading and probe interpolation, our method achieves state-of-the-art quality on reflective benchmarks. Overall, our results demonstrate that carefully designed explicit representations can serve as powerful and efficient alternatives to neural appearance models, enabling high-quality view-dependent effects, encouraging interpretability and practical rendering performance.
\clearpage
{
    \small
    \bibliographystyle{ieeenat_fullname}
    \bibliography{main}
}

\clearpage
\setcounter{page}{1}
\maketitlesupplementary\

\section*{Supplementary Overview}
In Section~\ref{sec:def_additional} we expand our deferred rendering formulation, explaining probe querying, near- and far-field illumination, and the role of roughness in controlling the Spherical Voronoi temperature. 
We then describe in Section~\ref{sec:speedup} our acceleration scheme for Spherical Voronoi evaluation, detailing the cubemap-based partitioning and truncated softmax used to reduce computational cost. 
Next, in Section~\ref{sec:tradeoff}~we present an analysis of the quality–efficiency tradeoff associated with varying the number of interpolated probes and sites. 
Section~\ref{sec:envmap} evaluates all bases in a controlled environment map fitting setting, isolating the contribution of SV from the complexity of a full 3DGS pipeline.
Section~\ref{sec:visuals}~provides additional qualitative visualizations: the adaptive SV decomposition learned when fitting environment maps, the effect of the temperature parameter $\tau$ on reflectance sharpness, and the optimization behavior of learnable light probes during training.
We also clarify the evaluation protocol adopted to ensure fair comparisons across datasets and prior work in Section~\ref{sec:evaluation}. 
Finally, in Section~\ref{sec:additional_results} we include additional quantitative results, including SV integration into other baselines and full  per-scene metrics across all datasets.

\section{Additional Details on Deferred Rendering}
\label{sec:def_additional}
In this section, we expand the reflection formulation introduced in the main paper  and provide a derivation of the near-field term $C_n$, the far field term $C_f$ and the spatial blending factor $\alpha$ appearing in Equation (\ref{eq:c_spec}). 

\subsection{Geometry Pass}
Following the deferred strategy illustrated in Figure \ref{fig:def_pipeline}, we first rasterize all 2DGS primitives once to produce the per-pixel attributes required for shading. For each pixel $(u,v)$, the geometry pass outputs:
\begin{itemize}
    \item world position $P(u,v) \in \mathbb{R}^3$,
    \item surface normal $N(u,v) \in \mathbb{S}^2$,
    \item diffuse color $D(u,v) \in \mathbb{R}^3$,
    \item roughness $R(u,v) \in [0,1]$.
\end{itemize}

These attributes are produced by splatting the corresponding Gaussian parameters, using the same weighted blending used during accumulation in standard 2DGS rendering.

\subsection{Lighting Pass}
Given the geometry buffer, the lighting pass computes the final shaded color:
\begin{equation}
    C(u,v) = D(u,v) + C_{\text{spec}}(u,v),
\end{equation}
where $C_{\text{spec}}(u,v)$ is the specular term defined in Equation (\ref{eq:c_spec}). We report it here for brevity:
\begin{equation}
    C_{\text{spec}}(u,v) = \alpha(u,v)C_n(u,v) + (1 - \alpha(u,v))C_f(u,v).
\end{equation}

Both illumination terms depend on the reflected view direction:
\begin{equation}
    \omega_r(u,v) = 2(\omega \cdot N(u,v))N(u,v) - \omega.
\end{equation}
In the following sections we describe how each component is computed.

\subsection{Near-field Illumination}
Near-field reflections are modeled by a set of learnable light probes
distributed in 3D space. Each probe $i$ has:
\begin{itemize}
    \item Position $p_i \in \mathbb{R}^3$,
    \item Spherical Voronoi function $f_i(\omega)$,
    \item Blend parameter $\alpha_i \in [0, 1]$.
\end{itemize}

\paragraph{Probe selection} We identify the probes that are most relevant for shading a given pixel $(u,v)$. We select the $k$ closest probes using Euclidean nearest-neighbor search:
\begin{equation}
    \mathcal{N}(u,v) = k\text{NN}(P(u,v)).
\end{equation}
This ensures that each pixel only interacts with probes that lie within a meaningful spatial neighborhood, reflecting the intuition that local geometry affects appearance only within a limited radius.

\paragraph{Distance-based weighting}
Not all selected probes contribute equally. Probes closer to the shading point should have a stronger influence than those further away. We therefore assign each probe a weight inversely proportional to its distance:
\begin{equation}
    w_i(u,v) = \frac{1}{\| P(u,v) - p_i \| + \epsilon},
\end{equation}
which is normalized as:
\begin{equation}
    \tilde{w}(u,v) = \frac{w_i(u,v)}
                            {\sum_{j \in \mathcal{N}(u,v)}w_j(u,v)}.
\end{equation}

The resulting weights vary smoothly throughout the image, producing spatially consistent transitions between lighting regions.

\paragraph{Evaluating and aggregating near-field illumination}
Each probe stores a SV representation of the reflected radiance at its location, which we query in the reflected direction $\omega_r(u,v)$ to obtain a directional estimate of the local illumination. The response of the selected probes is then combined using the normalized spatial weights $\tilde{w}_i(u,v)$. Formally, the near-field term is:
\begin{equation}
    C_n(u,v) = \sum_{i \in \mathcal{N}(u,v)} \tilde{w}_i(u,v)f_i(\omega_r(u,v)).
\end{equation}
This operation blends the directional information encoded at each probe according to its proximity to the shading point. 

\subsection{Far-field Illumination }
While near-field probes capture illumination effects produced by nearby geometry, many reflective surfaces are dominated by light coming from far more distant parts of the environment, such as skylight, outdoor structures, windows, or large surrounding objects. To represent this global component, we use a learnable environment cubemap.
Once the reflected direction $\omega_r(u,v)$ has been computed, obtaining the far-field illumination is straightforward: we simply sample the cubemap at that direction,
\begin{equation}
    C_f(u,v) = \text{cubemap}(\omega_r(u,v)).
\end{equation}

\subsection{Blending Factor }
The relative influence between near-field and far-field illumination must vary across the scene. Close to complex geometry, reflections are more sensitive to local variations and should rely predominantly on the probes. In open regions, distant illumination captured by the cubemap becomes more important. To achieve this adaptive behavior, each probe carries a learned blend weight $\alpha_i$, which express how strongly that probe favors near-field effects. At shading time, the per-pixel blend factor $\alpha(u,v)$ is obtained by interpolating these probe parameters using the same spatial weights $\tilde{w}_i(u,v)$ computed above:

\begin{equation}
    \alpha(u,v) = \sum_{i \in \mathcal{N}(u,v)}\tilde{w}_i(u,v)\alpha_i.
\end{equation}
This produces a smooth spatial field that naturally transitions between locally dominated and globally dominated reflection regimes.

\subsection{Roughness and Temperature}
The sharpness of the Spherical Voronoi representation is controlled by the temperature parameter $\tau$. Each Gaussian primitive stores its own roughness value $R$, which is splatted into a per-pixel roughness map $\mathbf{R}(\mathbf{x})$ during the geometry pass.
Roughness determines how sharp or smooth the directional function should be. To couple the reflectance model with Spherical Voronoi expressivity, we linearly map roughness to temperature:
\begin{equation}
    \tau(u,v) = (1 - R(u,v))\tau_{\text{max}} + R(u,v)\tau_{\text{min}},
\end{equation}
where we empirically set $\tau_{\text{min}} = 0.2$ and $\tau_{\text{max}} = 1500$. As a result:
\begin{itemize}
    \item \textit{low roughness} ($R \approx 0$) produces crisp, mirror-like reflections through large temperatures,
    \item \textit{high roughness} ($R \approx 1$) yields broad, diffuse responses via small temperatures.
\end{itemize}

% \subsection{Per-site Temperature}
% In the main paper (Sec.~\ref{sec:sv}), we introduce the soft Voronoi formulation with a single global temperature~$\tau$, which controls the overall smoothness of the partition. 
% For the \emph{radiance} model, we adopt a slightly more general variant where each site~$s_k$ carries its own implicit temperature, defined as
% \begin{equation}
%     \tau_k = \| s_k \|.
% \end{equation}
% In this parameterization, the direction $\hat{s}_k = s_k / \|s_k\|$ determines the site position on the sphere, while its magnitude acts as a learnable sharpness term. 
% This coupling allows the network to jointly adjust the position and softness of each lobe without introducing an extra scalar parameter, resulting in a compact and expressive representation. 
% Empirically, the model shows smoother optimization and slightly better radiance reconstruction in regions where the required sharpness varies across directions. 
% Conceptually, this remains the same differentiable soft Voronoi formulation described in the paper, but with site-dependent temperatures that adapt automatically during training. 
% For the \emph{reflection} model, $\tau$ remains explicitly controlled by surface roughness as discussed above.

\section{Speeding up Spherical Voronoi}
\label{sec:speedup}
A naïve evaluation of our Spherical Voronoi representation requires evaluating all $K$ sites for every queried direction $\omega$. While this is not a problem in the view-direction parameterization context, when modeling reflections which require a large number of sites, it becomes impractical. 
Concretely, for a function 
\begin{equation}
    f_{\mathrm{SV}}(\omega; \tau, s, c)
    = \sum_{k=1}^{K} w_k(\omega; \tau)\, c_k,
\end{equation}

where $\tau$ is the temperature controlling the sharpness of the partition, $s=\{s_k\}$ are the unit vectors defining the angular locations of the SV sites, and $c=\{c_k\}$ are their associated radiance values, with 
\begin{equation}
    w_k(\omega; \tau)
    = \frac{\exp(\tau\, s_k \cdot \omega)}
           {\sum_{k'=1}^{K} \exp(\tau\, s_{k'} \cdot \omega)},
\end{equation}
the cost of a single evaluation scales linearly with the number of sites $K$. When using thousands of sites per function, this becomes a major bottleneck during rendering, especially when evaluating directional appearance at every pixel. To mitigate this cost, we introduce a simple acceleration scheme that exploits the fact that \textit{only a small subset of sites is relevant for any given direction}. Our key idea is to partition the unit sphere using a low-resolution cubemap, and to pre-assign to each texel a fixed set of \textit{candidate} sites. At runtime, the softmax is restricted to this candidate set instead of all sites.

\subsection{Cubemap-Based Partition of the Sphere}
We discretize the sphere using a cubemap of fixed resolution. Each texel $t$ in the cubemap is associated with a direction $\hat{\omega}_t$ corresponding to its center. In a preprocessing step, for each texel $t$ we select a small subset of sites $\mathcal{S}(t) \subset \{ 1, ..., K \}$ that are the most relevant for directions around $\hat{\omega}_t$. A natural choice is to pick the sites with highest similarity to $\hat{\omega}_t$, \ie:
\begin{equation}
    \mathcal{S}(t) = \text{TopK}_k(  s_k\cdot \hat{\omega}_t),
\end{equation}
where TopK denotes the set of indices of the $k$ closest sites. This yields a lookup table that maps each cubemap texel to a much smaller set of candidate sites. Since the Spherical Voronoi sites are optimized jointly with the rest of the model, the cubemap-to-site assignment $\mathcal{S}(t)$ becomes outdated over the course of training. To account for this, we periodically recompute it. In all our experiments, we rebuild the assignment every 500 optimization steps, which we found to be a good compromise between accuracy of the approximation and preprocessing overhead. 
%\dor{What's $s_k$?}

\subsection{Softmax Approximation}
At rendering time, given a direction $\omega_r$, we determine the cubemap texel $t(\omega_r)$ that contains $\omega_r$. Instead of evaluating the SV weights over all sites, we restrict the computation to the precomputed candidate set. The accelerated SV evaluation becomes:
\begin{equation}
    f_{\mathrm{SV}}^{\text{fast}}(\omega_r)
        = \sum_{k \in \mathcal{S}(t(\omega_r))}
        \tilde{w}_k(\omega_r; \tau)\, c_k,
\end{equation}
where the truncated weights are defined as:
\begin{equation}
\tilde{w}_k(\omega_r; \tau)
= \frac{\exp(\tau\, s_k \cdot \omega_r)}
       {\sum_{k' \in \mathcal{S}(t(\omega_r))} 
              \exp(\tau\, s_{k'} \cdot \omega_r)}.
\end{equation}

In other words, we approximate the full softmax over $K$ sites with a local softmax over a small, direction-dependent subset of sites. Intuitively, for sufficiently fine cubemap resolution and moderately large candidate set size, the omitted sites have negligible contribution to the softmax, as their dot product $s_k \cdot \omega_r$ is much smaller. This reduces the per-direction complexity from $\mathcal{O}(K)$ to $\mathcal{O}(|\mathcal{S}(t)|)$.

\section{Quality-Efficiency Tradeoff}
\label{sec:tradeoff}

\paragraph{Number of SV sites (radiance modeling)}
We use 8 sites, matching the parameter count of order-3 spherical 
harmonics for a fair comparison. Table~\ref{tab:ab2} shows results 
varying the number of sites from 1 to 16: performance saturates 
around 8--12 sites with diminishing returns beyond, while inference 
cost increases linearly.

\begin{table}[t]
\centering
\caption{Effect of the number of sites on quality and efficiency, 
averaged over \textit{bonsai} and \textit{garden} scenes of 
\textit{Mip-NeRF 360}.}
\label{tab:ab2}
\resizebox{0.85\columnwidth}{!}{%
\begin{tabular}{ccccccc}
\multicolumn{1}{l}{} & \#sites & PSNR $\uparrow$ & SSIM $\uparrow$ & LPIPS $\downarrow$ & FPS $\uparrow$ & Train Time (min) $\downarrow$ \\ \hline
                     & 1  & 29.64 & 0.905 & 0.186 & 295 & 21 \\
                     & 2  & 30.74 & 0.912 & 0.178 & 268 & 22 \\
                     & 4  & 31.04 & 0.914 & 0.177 & 226 & 25 \\
                     & \cellcolor[HTML]{FF9998}8  & \cellcolor[HTML]{FF9998}31.21 & \cellcolor[HTML]{FF9998}0.915 & \cellcolor[HTML]{FF9998}0.175 & \cellcolor[HTML]{FF9998}173 & \cellcolor[HTML]{FF9998}32 \\
                     & 12 & 31.23 & 0.915 & 0.173 & 138 & 39 \\
\multirow{-6}{*}{SV} & 16 & 31.28 & 0.915 & 0.173 & 116 & 47 \\ \hline
SB                   & -  & 30.71 & 0.913 & 0.177 & 148 & 28 \\ \hline
SG                   & -  & 30.37 & 0.911 & 0.176 & 104 & 31 \\ \hline
SH                   & -  & 30.39 & 0.912 & 0.176 & 107 & 30
\end{tabular}%
}
\end{table}

\paragraph{Probe and site capacity (reflection modeling)}
\begin{figure}[t]
\centering
\includegraphics[width=0.9\columnwidth]{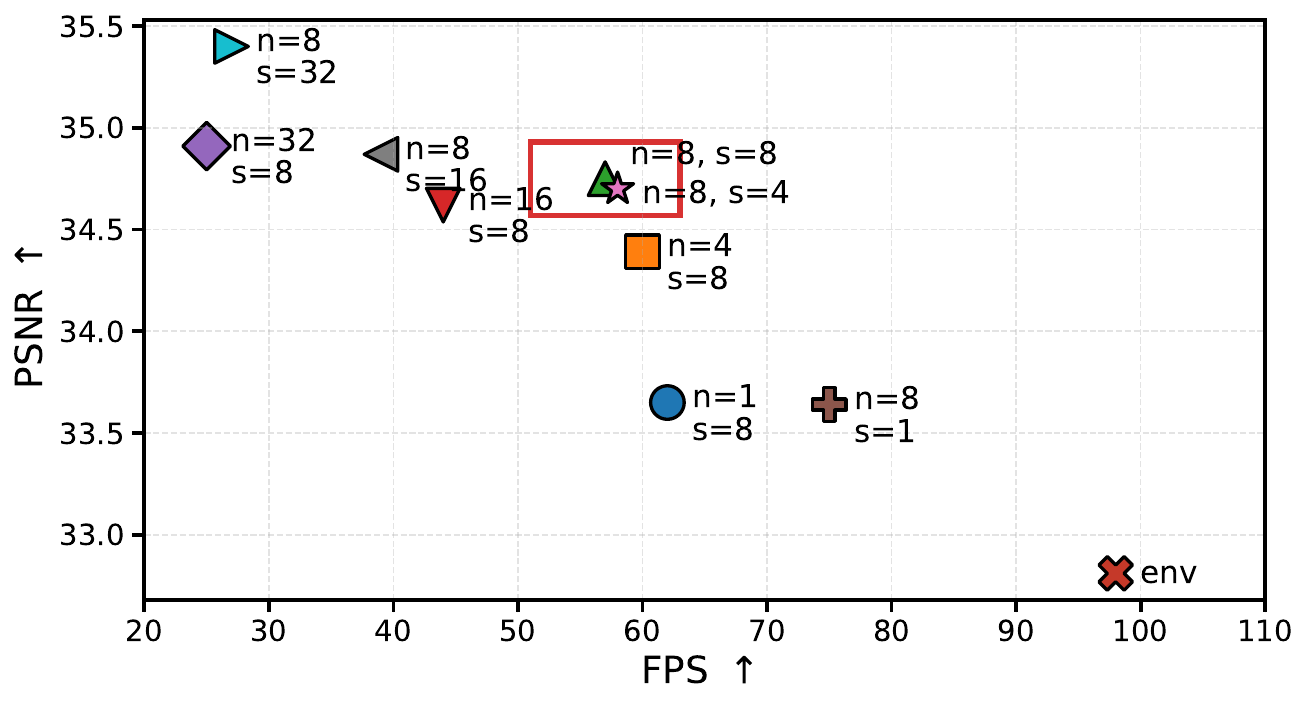}

\vspace{-0.5em}
\caption{\textbf{Quality–Efficiency Tradeoff} - Increasing the kernel capacity ($|\mathcal{N}(K)|$ and $|\mathcal{S}(K)|$; abbreviated as $n$ and $s$ in the plot) leads to higher rendering quality but lower inference speed, highlighting the fundamental tradeoff between accuracy and efficiency. Results are reported for the \textit{coffee} scene of the Ref-NeRF dataset. \textit{Env} denotes the configuration without probes.}

\label{fig:tradeoff}
\end{figure}

We evaluate the effect of kernel capacity by varying the 
cardinalities $|S(K)|$ and $|N(K)|$. As shown in 
Figure~\ref{fig:tradeoff}, increasing these values improves 
reconstruction quality but reduces inference speed, exhibiting a 
clear quality--efficiency tradeoff. Notably, setting $|N(K)|=32$ 
yields a small quality gain but nearly halves the FPS, showing 
rapidly diminishing returns. We therefore adopt $|S(K)|=8$ and 
$|N(K)|=8$ as a balanced configuration that preserves most of the 
quality benefits while maintaining real-time performance.

\section{Environment Map Fitting}
\label{sec:envmap}

\begin{figure*}[t]
    \centering
    \includegraphics[width=\linewidth]{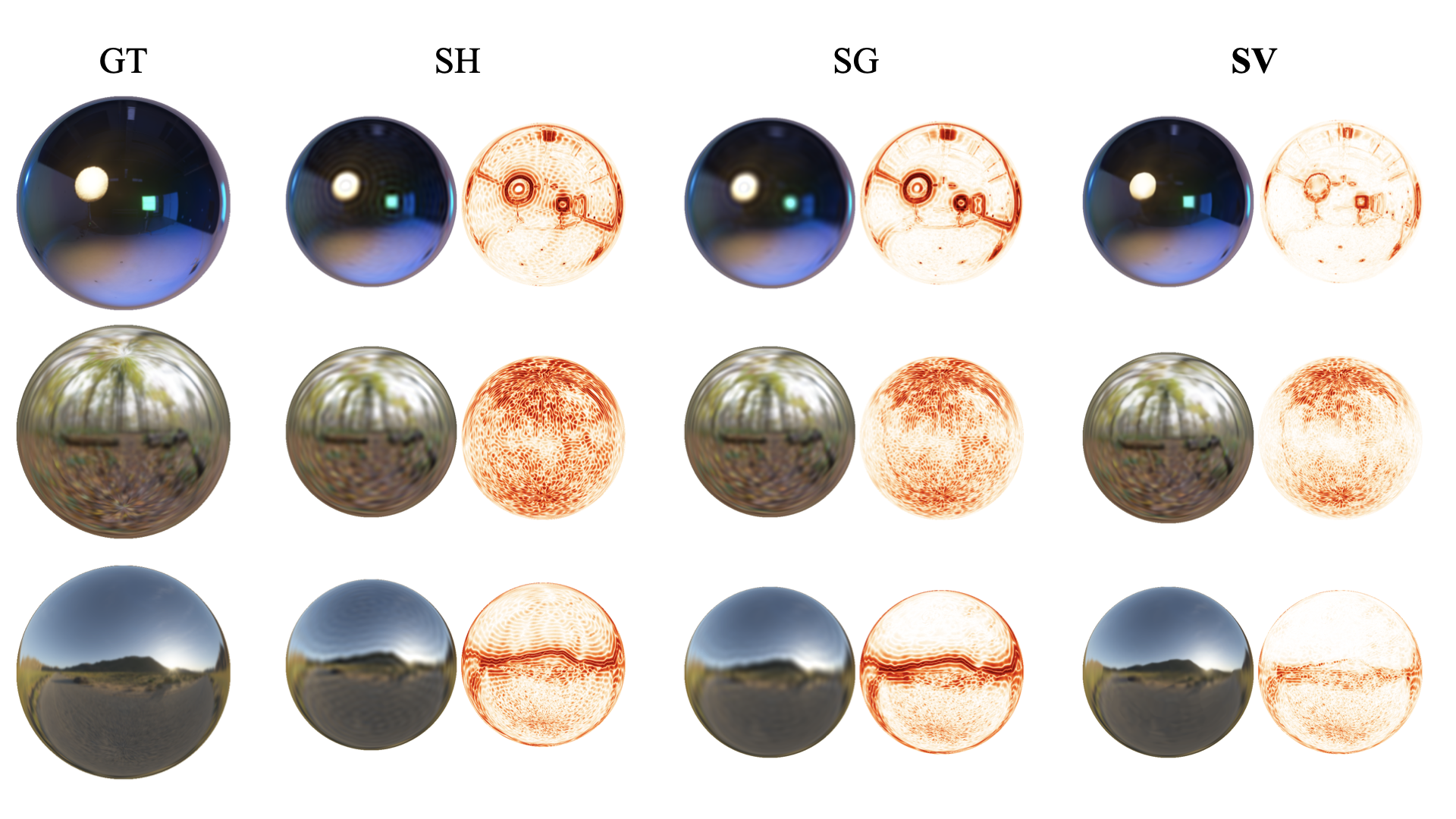}
    \caption{
        \textbf{Environment map fitting.}
        For each method we show the rendered reflective sphere (left) and the 
        corresponding error map (right). All methods are optimized under the 
        same parameter budget.
    }
    \label{fig:envmap}
\end{figure*}

To isolate the contribution of the SV representation from the complexity of a full 3DGS pipeline, we evaluate all bases in a controlled setting: fitting a known environment map onto a reflective sphere. We select environment maps that contain both high-frequency content (trees, road, grass) and low-frequency regions (sky), providing a challenging yet balanced test case. We fix the parameter budget 
across all methods and report PSNR, SSIM, and LPIPS at convergence in  Table~\ref{tab:envmap} and Figure~\ref{fig:envmap}. SV consistently 
outperforms SH and SG across all scenes and metrics, with notably cleaner error maps.

\begin{table}[t]
\centering
\caption{\textbf{Environment map fitting.} PSNR, SSIM, and LPIPS at convergence under a fixed parameter budget.}
\label{tab:envmap}
\resizebox{0.75\columnwidth}{!}{%
\begin{tabular}{cc|ccc}
Scene                    & Method      & PSNR $\uparrow$                         & SSIM$\uparrow$                          & LPIPS$\downarrow$                         \\ \hline
                         & \textbf{SV} & \cellcolor[HTML]{FF9998}38.58 & \cellcolor[HTML]{FF9998}0.978 & \cellcolor[HTML]{FF9998}0.087 \\
                         & SG          & 32.66                         & 0.940                         & 0.180                         \\
\multirow{-3}{*}{studio} & SH          & 31.37                         & 0.946                         & 0.185                         \\ \hline
                         & \textbf{SV} & \cellcolor[HTML]{FF9998}36.42 & \cellcolor[HTML]{FF9998}0.944 & \cellcolor[HTML]{FF9998}0.087 \\
                         & SG          & 34.17                         & 0.917                         & 0.136                         \\
\multirow{-3}{*}{forest} & SH          & 31.78                         & 0.872                         & 0.179                         \\ \hline
                         & \textbf{SV} & \cellcolor[HTML]{FF9998}40.36 & \cellcolor[HTML]{FF9998}0.969 & \cellcolor[HTML]{FF9998}0.106 \\
                         & SG          & 34.03                         & 0.932                         & 0.174                         \\
\multirow{-3}{*}{road}    & SH          & 32.51                         & 0.930                         & 0.190                        
\end{tabular}%
}
\end{table}

\section{Visualization of SV Behavior and Probe Dynamics}
\label{sec:visuals}

\begin{figure*}
    \centering
    \includegraphics[width=\linewidth]{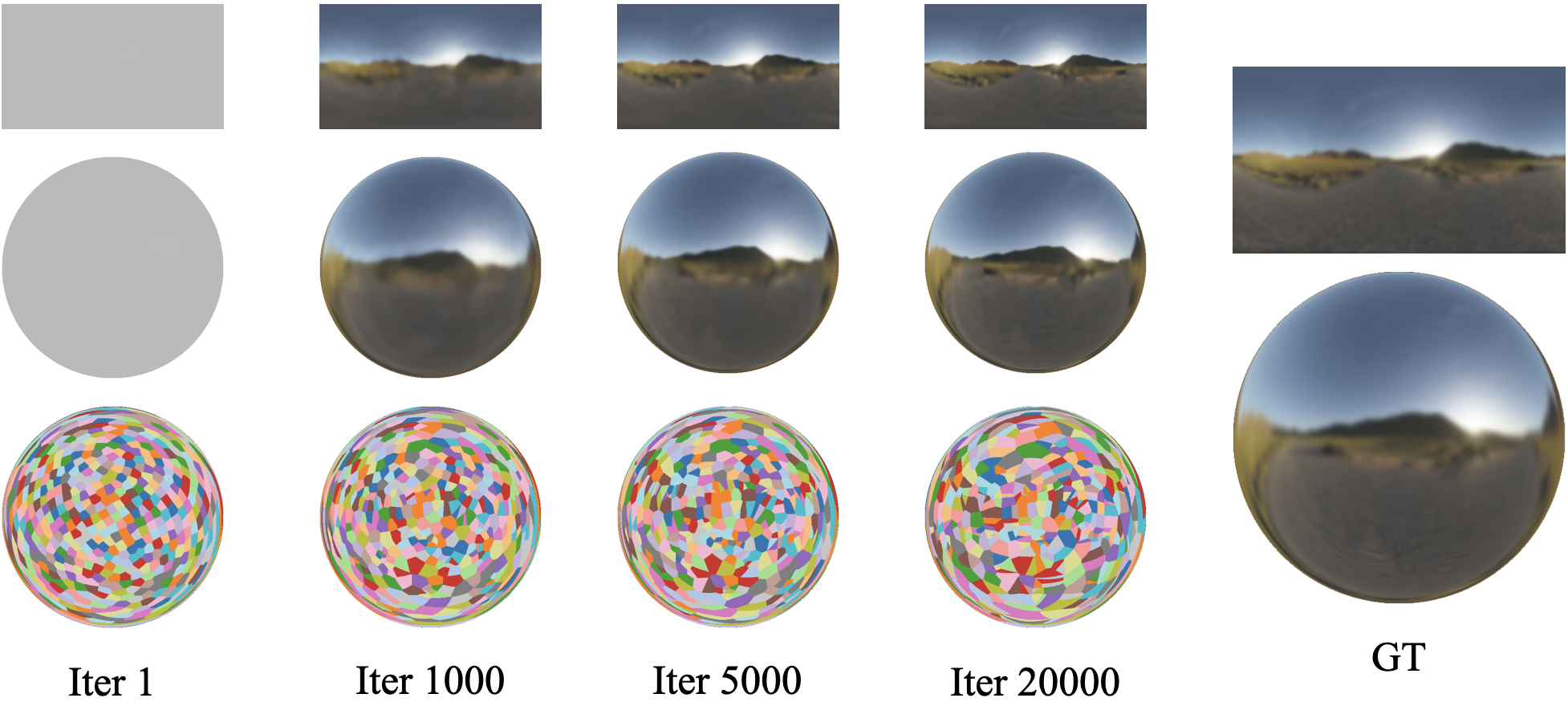}
    \caption{\textbf{SV evolution during training (3D)} - Top: predicted environment map in lat-long format. Middle: its rendering on a reflective sphere. Bottom: corresponding SV tessellation (colored by random site IDs). Across training iterations, the SV sites reorganize into a structured decomposition of the sphere. Ground truth is shown on the right.}
    \label{fig:sv_learning}
\end{figure*}

\begin{figure*}
    \centering
    \includegraphics[width=0.9\linewidth]{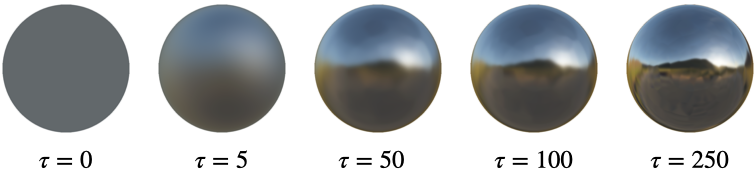}
    \caption{\textbf{Effect of the SV temperature $\tau$ (3D)} - Increasing $\tau$ progressively sharpens the SV lobes, transitioning from an almost uniform shading ($\tau = 0$) to increasingly crisp, mirror-like reflections. Renderings shown for $\tau \in \{0, 5, 50, 100, 250\}$}
    \label{fig:sv_temp}
\end{figure*}

\begin{figure*}
    \centering
    \includegraphics[width=0.9\linewidth]{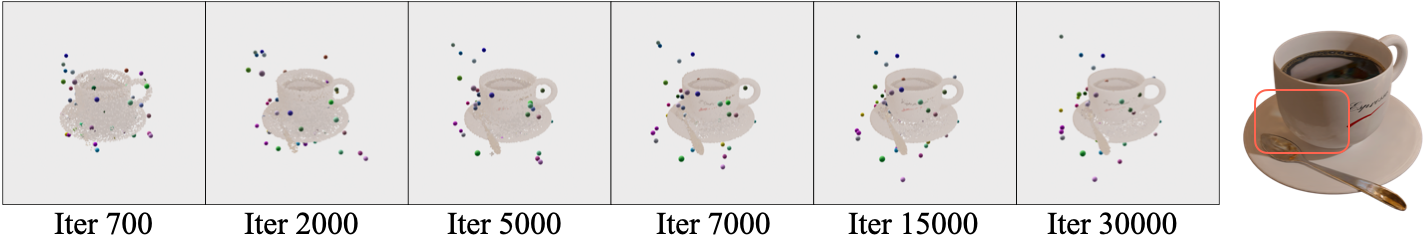}
    \caption{\textbf{Evolution of learnable light probes during training} -  The probes are initialized at random positions and gradually migrate toward regions exhibiting strong near-field specular interactions. Throughout optimization, they consistently cluster around the reflective side of the object (e.g., the spoon-facing region in the Coffee scene of the Ref-NeRF dataset), which corresponds to an area rich in local specular highlights and interreflections.}
    \label{fig:probe_learning}
\end{figure*}

In this section, we provide qualitative visualizations that further illustrate the behavior of our SV representation and the dynamics of learnable light probes. Figure~\ref{fig:sv_learning}~shows the adaptive decomposition of the spherical domain when learning an environment map. Figure~\ref{fig:sv_temp} visualizes how increasing $\tau$ sharpens SV lobes and enhances specular detail. Figure~\ref{fig:probe_learning}~illustrates the optimization trajectory of the light probes.

\section{Clarifications on Evaluation}
\label{sec:evaluation}
To ensure a fair and consistent comparison across methods, we reviewed the evaluation protocols commonly used for the Mip-NeRF~360 and Ref-Real datasets. We found that some prior works rely on testing setups that deviate from the dataset guidelines, which can lead to inflated performance metrics. For Mip-NeRF~360, Beta Splatting \cite{dbs} evaluates models on images that are downsampled on-the-fly from the high-resolution originals (using the \texttt{--r} flag in 3DGS codebase). This produces smoother inputs that are easier to fit, increasing PSNR by approximately 0.5\,dB on average compared to using the official downsampled images provided in the dataset.
Additionally, Beta Splatting employs a checkpoint-selection mechanism: starting from iteration 15k, the model is evaluated on the test set every 500 steps, and the best checkpoint is reported as the final result. While practical, this introduces a form of test-set selection that is not fully aligned with standard evaluation practice.
A similar issue appears in the Ref-Real results reported by Ref-GS \cite{refgs}, where non-standard downsampling choices also deviate from the dataset's recommended evaluation setup.
For a fully fair and reproducible evaluation, we strictly follow the dataset recommendations. In Mip-NeRF~360, we use \texttt{images\_2} for indoor scenes and \texttt{images\_4} for outdoor scenes. For Ref-Real, we use \texttt{images\_4} for \emph{garden spheres} and \emph{toy car}, and \texttt{images\_8} for \emph{sedan}. All results in this paper follow these settings.

\section{Additional Results}
\label{sec:additional_results}
\subsection{Modeling Radiance}
\begin{table*}[ht]
\centering
\caption{\textbf{Radiance Modeling with SV Voronoi} - 
Applying SV to existing Gaussian Splatting baselines yields consistent improvements across datasets.}
\label{tab:additional_radiance}
\resizebox{0.75\linewidth}{!}{%
\begin{tabular}{clccc|ccc}
\multicolumn{1}{l}{}                                                                 &                            & \multicolumn{3}{c|}{Vanilla}                                          & \multicolumn{3}{c}{Ours (SV)}                                                                 \\
\multicolumn{1}{l}{}                                                                 & \multicolumn{1}{c}{Method} & PSNR $\uparrow$  & SSIM $\uparrow$                          & LPIPS $\downarrow$                         & PSNR $\uparrow$                          & SSIM $\uparrow$                          & LPIPS $\downarrow$                         \\ \hline
                                                                                     & \textit{3DGS-mcmc}         & 27.92 & 0.833                         & 0.234                         & \cellcolor[HTML]{FF9998}28.51 & \cellcolor[HTML]{FF9998}0.837 & \cellcolor[HTML]{FF9998}0.220 \\
                                                                                     & \textit{2DGS}              & 26.86 & 0.798                         & 0.301                         & \cellcolor[HTML]{FF9998}27.70 & \cellcolor[HTML]{FF9998}0.807 & \cellcolor[HTML]{FF9998}0.279 \\
\multirow{-3}{*}{\textit{\begin{tabular}[c]{@{}c@{}}Mip-NeRF \\ 360\end{tabular}}}   & \textit{Beta-Splatting}    & 28.12 & 0.831                         & 0.238                         & \cellcolor[HTML]{FF9998}28.71 & \cellcolor[HTML]{FF9998}0.837 & \cellcolor[HTML]{FF9998}0.228 \\ \hline
                                                                                     & \textit{3DGS-mcmc}         & 33.77 & 0.972                         & 0.036                         & \cellcolor[HTML]{FF9998}34.35 & \cellcolor[HTML]{FF9998}0.973 & \cellcolor[HTML]{FF9998}0.034 \\
                                                                                     & \textit{2DGS}              & 33.17 & 0.968                         & 0.041                         & \cellcolor[HTML]{FF9998}33.56 & \cellcolor[HTML]{FF9998}0.969 & \cellcolor[HTML]{FF9998}0.039 \\
\multirow{-3}{*}{\textit{\begin{tabular}[c]{@{}c@{}}NeRF-\\ Synthetic\end{tabular}}} & \textit{Beta-Splatting}    & 34.10 & 0.971                         & 0.034                         & \cellcolor[HTML]{FF9998}34.58 & \cellcolor[HTML]{FF9998}0.973 & \cellcolor[HTML]{FF9998}0.032 \\ \hline
                                                                                     & \textit{3DGS-mcmc}         & 24.24 & 0.863                         & 0.190                         & \cellcolor[HTML]{FF9998}24.67 & \cellcolor[HTML]{FF9998}0.867 & \cellcolor[HTML]{FF9998}0.186 \\
\multirow{-2}{*}{\textit{Tanks\&Temples}}                                            & \textit{Beta-Splatting}    & 24.54 & 0.871                         & 0.171                         & \cellcolor[HTML]{FF9998}25.00 & \cellcolor[HTML]{FF9998}0.874 & \cellcolor[HTML]{FF9998}0.170 \\ \hline
                                                                                     & \textit{3DGS-mcmc}         & 29.55 & 0.901                         & 0.320                         & \cellcolor[HTML]{FF9998}30.04 & \cellcolor[HTML]{FF9998}0.905 & \cellcolor[HTML]{FF9998}0.315 \\
\multirow{-2}{*}{\textit{DeepBlending}}                                              & \textit{Beta-Splatting}    & 29.56 & 0.907                         & 0.316                         & \cellcolor[HTML]{FF9998}30.63 & \cellcolor[HTML]{FF9998}0.917 & \cellcolor[HTML]{FF9998}0.298
\end{tabular}%
}
\end{table*}
SV can be seamlessly integrated into other Gaussian Splatting pipelines. In Table~\ref{tab:additional_radiance}, we report the improvements obtained when augmenting two popular baselines—3DGS-MCMC~\cite{3dgs-mcmc} and 2DGS~\cite{2dsplat}—with our SV modeling. The consistent gains across both methods demonstrate that SV serves as a general and effective mechanism for enhancing radiance modeling in Gaussian-based representations.

\subsection{Per-scene Results}
\begin{table}[ht]
\centering
\caption{\textbf{Per-scene radiance quality} -
Rendering metrics reported per scene across all evaluated datasets.}
\label{tab:per_scene_radiance}
\resizebox{0.7\columnwidth}{!}{%
\begin{tabular}{clccc}
\multicolumn{1}{l}{}                                                                & Scene     & PSNR $\uparrow$  & SSIM $\uparrow$  & LPIPS $\downarrow$ \\ \hline
\multirow{10}{*}{\textit{\begin{tabular}[c]{@{}c@{}}Mip-NeRF \\ 360\end{tabular}}}  & bicycle   & 25.75 & 0.795 & 0.196 \\
                                                                                    & bonsai    & 34.87 & 0.957 & 0.231 \\
                                                                                    & counter   & 31.04 & 0.930 & 0.225 \\
                                                                                    & flowers   & 22.18 & 0.639 & 0.337 \\
                                                                                    & garden    & 27.92 & 0.876 & 0.114 \\
                                                                                    & kitchen   & 32.98 & 0.937 & 0.147 \\
                                                                                    & room      & 33.26 & 0.937 & 0.257 \\
                                                                                    & stump     & 27.32 & 0.806 & 0.211 \\
                                                                                    & treehill  & 23.08 & 0.656 & 0.333 \\
                                                                                    & mean      & 28.71 & 0.837 & 0.228 \\ \hline
\multirow{9}{*}{\textit{\begin{tabular}[c]{@{}c@{}}NeRF-\\ Synthetic\end{tabular}}} & chair     & 37.18 & 0.990 & 0.012 \\
                                                                                    & drums     & 27.05 & 0.959 & 0.037 \\
                                                                                    & ficus     & 36.93 & 0.991 & 0.009 \\
                                                                                    & hotdog    & 38.49 & 0.988 & 0.020 \\
                                                                                    & lego      & 37.15 & 0.986 & 0.015 \\
                                                                                    & materials & 30.98 & 0.967 & 0.036 \\
                                                                                    & mic       & 37.39 & 0.994 & 0.006 \\
                                                                                    & ship      & 31.49 & 0.911 & 0.117 \\
                                                                                    & mean      & 34.58 & 0.973 & 0.032 \\ \hline
\multirow{3}{*}{\textit{\begin{tabular}[c]{@{}c@{}}Tanks\&\\ Temples\end{tabular}}} & train     & 23.25 & 0.845 & 0.214 \\
                                                                                    & truck     & 26.76 & 0.904 & 0.125 \\
                                                                                    & mean      & 25.00 & 0.874 & 0.170 \\ \hline
\multirow{3}{*}{\textit{\begin{tabular}[c]{@{}c@{}}Deep\\ Blending\end{tabular}}}   & drjohnson & 29.96 & 0.914 & 0.304 \\
                                                                                    & playroom  & 31.30 & 0.920 & 0.292 \\
                                                                                    & mean      & 30.63 & 0.917 & 0.298
\end{tabular}%
}
\end{table}
\begin{table}[ht]
\centering
\caption{\textbf{Per-scene reflection quality} - 
Rendering metrics reported per scene across all evaluated datasets.}
\label{tab:per-scene_reflections}
\resizebox{0.75\columnwidth}{!}{%
\begin{tabular}{llccc}
\multicolumn{1}{l}{}                                                        & \multicolumn{1}{l}{Scene} & PSNR $\uparrow$  & SSIM $\uparrow$  & LPIPS $\downarrow$ \\ \hline
\multirow{7}{*}{Ref-NeRF}                                                   & ball                      & 39.40 & 0.987 & 0.082 \\
                                                                            & car                       & 31.26 & 0.968 & 0.029 \\
                                                                            & coffee                    & 34.91 & 0.973 & 0.082 \\
                                                                            & helmet                    & 35.63 & 0.983 & 0.031 \\
                                                                            & teapot                    & 47.59 & 0.997 & 0.007 \\
                                                                            & toaster                   & 27.73 & 0.950 & 0.070 \\
                                                                            & mean                      & 36.09 & 0.976 & 0.050 \\ \hline
\multirow{7}{*}{\begin{tabular}[c]{@{}c@{}}Glossy\\ Synthetic\end{tabular}} & bell                      & 32.22 & 0.963 & 0.044 \\
                                                                            & cat                       & 33.46 & 0.975 & 0.033 \\
                                                                            & luyu                      & 30.22 & 0.950 & 0.042 \\
                                                                            & potion                    & 33.74 & 0.961 & 0.061 \\
                                                                            & tbell                     & 30.90 & 0.966 & 0.050 \\
                                                                            & teapot                    & 27.25 & 0.954 & 0.048 \\
                                                                            & mean                      & 31.30 & 0.962 & 0.046 \\ \hline
\multirow{4}{*}{Ref-Real}                                                   & gardenspheres             & 21.97 & 0.578 & 0.275 \\
                                                                            & sedan                     & 25.94 & 0.757 & 0.215 \\
                                                                            & toycar                    & 23.83 & 0.641 & 0.243 \\
                                                                            & mean                      & 23.91 & 0.659 & 0.244
\end{tabular}%
}
\end{table}

We provide a detailed per-scene evaluation for all datasets used in the main paper. 
Table~\ref{tab:per_scene_radiance} reports results on radiance-dominated datasets 
(Mip-NeRF~360, NeRF-Synthetic, Tanks\&Temples, and Deep Blending), while 
Table~\ref{tab:per-scene_reflections} presents results for reflection-heavy datasets 
(Ref-NeRF, Glossy Synthetic, and Ref-Real).

\end{document}